%% file: BlueLM-GUI.tex
\documentclass[11pt, a4paper, copyright]{google}

\usepackage{tikz}
\usepackage{forest}
\usepackage{multicol}
\usetikzlibrary{positioning,arrows.meta,shapes.geometric,fit,backgrounds,calc,decorations.pathreplacing}

\forestset{
  leafstyle/.style={draw=green!50!black, rounded corners=2pt, fill=green!10, minimum height=20pt, inner xsep=6pt, align=center},
  reusestyle/.style={draw=gray, dashed, rounded corners=2pt, fill=gray!10, minimum height=20pt, inner xsep=6pt, align=center},
}

\usepackage[authoryear, sort&compress, round]{natbib}
\makeatletter
\renewcommand{\absfont}{\normalfont\linespread{1.2}\fontsize{11}{12}\selectfont}

\definecolor{zjublue}{RGB}{0,63,136}      
\definecolor{zjulightblue}{RGB}{0,159,227} 

\usepackage[colorlinks = true,
            linkcolor = zjublue,
            urlcolor  = zjulightblue,
            citecolor = zjublue,
            anchorcolor = zjublue]{hyperref}
\usepackage[misc]{ifsym}
\usepackage{minitoc}
\usepackage{cleveref}
\usepackage{hyperref}
\usepackage{subcaption}
\usepackage{booktabs}
\usepackage{fontawesome5}
\usepackage{amsmath}
\usepackage{mathtools}
\usepackage{amssymb}
\usepackage{graphicx}
\usepackage{algorithmic}
\usepackage{algorithm}
\usepackage{tcolorbox}
\usepackage{verbatim}
\usepackage{makecell}
\usepackage{array}
\usepackage{multirow}
\usepackage{xurl}
\usepackage{amsthm}
\usepackage{adjustbox}
\usepackage{caption}
\usepackage{xcolor}
\usepackage{wrapfig}
\usepackage{pifont}
\usepackage{enumitem}

\usepackage{colortbl}
\usepackage{xcolor}

\usepackage{titlesec}
\titlespacing*{\paragraph}{0pt}{0pt}{1em}

\definecolor{headerblue}{RGB}{180, 210, 240}
\definecolor{rowgray}{RGB}{248, 248, 248}
\definecolor{closedrow}{RGB}{255, 245, 220}
\definecolor{groupblue}{RGB}{220, 235, 250}
\definecolor{successgreen}{RGB}{46, 139, 87}
\definecolor{failred}{RGB}{200, 60, 60}
\definecolor{groupheader}{RGB}{230, 230, 230}
\definecolor{sectionblue}{RGB}{198, 217, 232}

\usepackage[titles]{tocloft}
\definecolor{tocsubsec}{HTML}{444444}
\definecolor{tocsubsubsec}{HTML}{666666}

\usepackage{soul}

\usepackage{listings}
\tcbuselibrary{listings,skins,breakable}

\lstdefinestyle{prompt}{
  basicstyle=\ttfamily\footnotesize,
  breaklines=true,
  breakatwhitespace=true,
  columns=fullflexible,
  keepspaces=true,
  showstringspaces=false,
  postbreak=\mbox{\textcolor{gray}{$\hookrightarrow$}\space}
}

\definecolor{accent1}{HTML}{2E6DA4}
\definecolor{accent2}{HTML}{D9534F}
\definecolor{accent3}{HTML}{5CB85C}
\definecolor{lightblue}{RGB}{173,216,230}
\definecolor{lightorange}{RGB}{255,213,170}
\definecolor{lightgreen}{RGB}{176,226,176}
\definecolor{lightyellow}{RGB}{255,255,204}
\definecolor{lightgray}{RGB}{220,220,220}
\definecolor{lightpurple}{RGB}{221,160,221}
\definecolor{lightred}{RGB}{255,182,193}
\definecolor{gray60}{gray}{0.6}
\definecolor{accent4}{HTML}{9B59B6}

\uselogo{}

\input{math_commands}

\usepackage{url}
\usepackage[utf8]{inputenc}
\usepackage{amsfonts}
\usepackage{nicefrac}
\usepackage{xspace}
\usepackage{comment}
\usepackage{minted}

\definecolor{highlightcolor}{RGB}{233,247,217}
\definecolor{highgrey}{RGB}{220,220,220}
\definecolor{lightblueB}{HTML}{88AFC9}

\newcommand{\OurModel}{BlueLM-GUI\xspace}          
\newcommand{\BenchBasic}{MobileGUI-VBench-Basic\xspace}  
\newcommand{\BenchPro}{MobileGUI-VBench\xspace}      

\newcommand{\ScreenEN}{ScreenSpot-v2\xspace}       
\newcommand{\ScreenZH}{ScreenSpot-ZH\xspace}
\AtBeginDocument{%
  \setlength\abovedisplayskip{0.6ex}
  \setlength\belowdisplayskip{0.6ex}}
\allowdisplaybreaks[4]
\makeatletter
\newcommand\figcaption{\def\@captype{figure}\caption}
\newcommand\tabcaption{\def\@captype{table}\caption}
\makeatother
\usepackage{upquote}
\usepackage{tcolorbox}
\newtcolorbox{codebox}[1][]{
    colback=green!10,
    colframe=green!50!black,
    boxrule=2pt,
    arc=5pt,
    left=10pt,
    right=10pt,
    top=10pt,
    bottom=10pt,
    title=#1 
}
\newcommand{\code}[1]{\texttt{#1}}

\makeatletter
\renewcommand{\maketitle}{\bgroup\setlength{\parindent}{0pt}
  \begin{adjustwidth}{0pt}{24pt}
    \begin{center}
      {\titlefont \@title\par}%
      \vskip11pt
      {\@author\par}%
      \vskip20pt%
    \end{center}
  \end{adjustwidth}
  \egroup
  {\abscontent}%
  \thispagestyle{firststyle}
}
\makeatother

\AtBeginDocument{%
  \setlength{\headsep}{28pt}
  \fancyhead[C]{\footerfont BlueLM-GUI Technical Report}%
  \fancypagestyle{firststyle}{%
    \fancyhead[L]{\hspace*{0pt}\raisebox{-0.3\height}{\includegraphics[height=30pt]{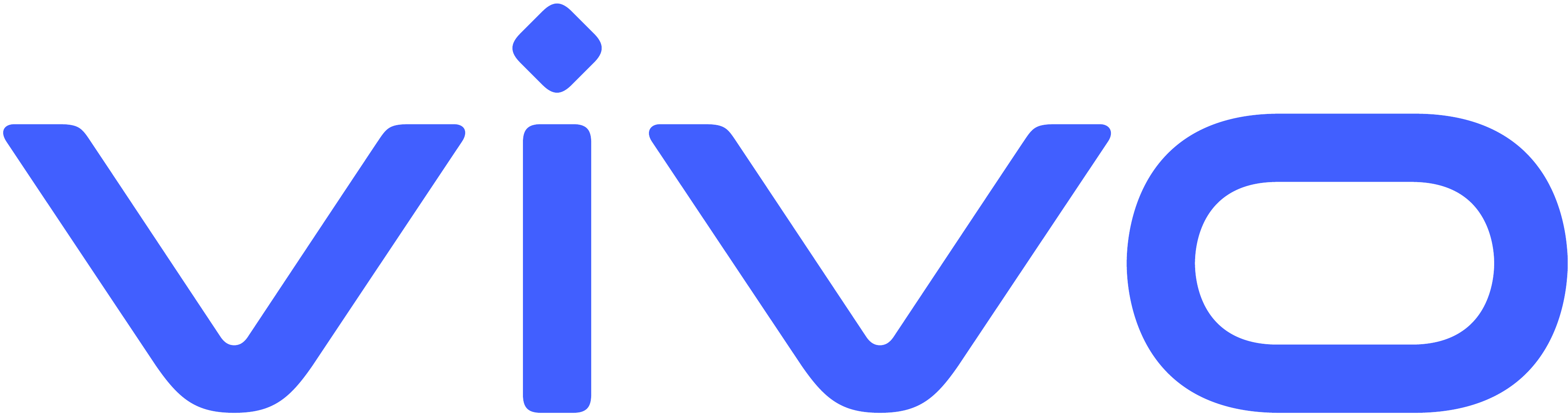}}}%
    \fancyhead[R]{\raisebox{-0.3\height}{\includegraphics[height=50pt, trim=0 0 0 0, clip]{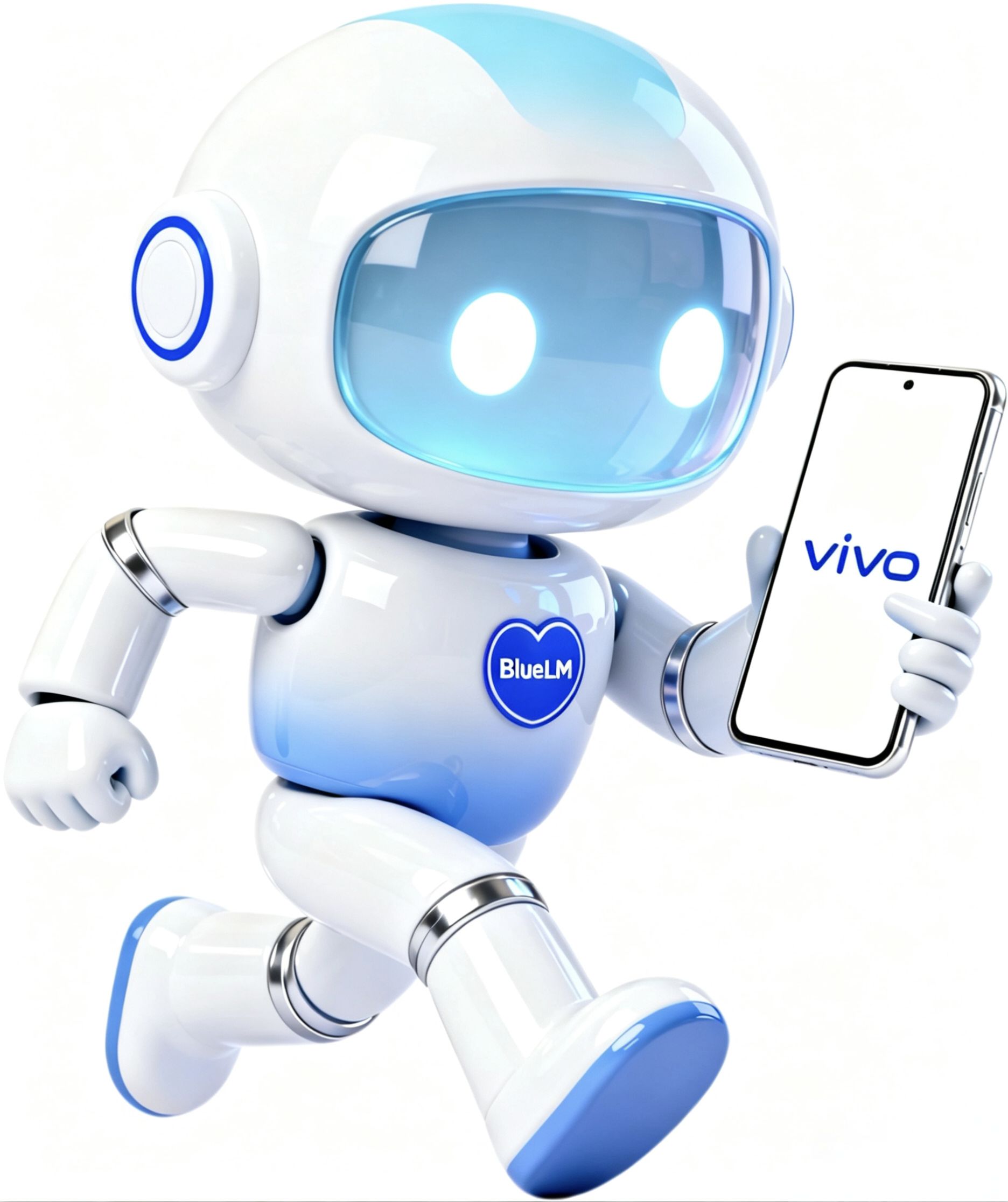}}\hspace*{2pt}}%

    \fancyhead[C]{}%
    \fancyfoot[L]{%
  \footerfont
  {\bfseries vivo AI Lab}. \quad
  \newline
}%
    \fancyfoot[R]{\footerfont\bfseries \thepage}%
    \fancyfoot[C]{\footerfont\bfseries\relax}%
  }%
  \fancyfoot[L]{\footerfont{           vivo AI Lab.}}%
  \fancyfoot[R]{\footerfont \thepage}%

}

\title{BlueLM-GUI Technical Report: \\ A Real-Device-Centric Flywheel for Self-Improving Mobile GUI Agents}

\author{%
  vivo AI Lab
  \par\vspace{10pt}
  \faGithub~\href{https://github.com/vivo-ai-lab/BlueLM-GUI}{\texttt{https://github.com/vivo-ai-lab/BlueLM-GUI}}  
}

\begin{abstract}
\textbf{Abstract.}
Mobile GUI agents are shifting from multi-module frameworks to native models trained end-to-end, yet industrial deployment faces three persistent gaps. Sandbox training produces a distribution mismatch with production environments; expensive real-device failures remain underutilized; and fixed benchmarks saturate, losing the power to guide iteration. We present \textbf{BlueLM-GUI}, a 35B-A3B mobile GUI agent built as a real-device-centric flywheel that closes these gaps through three principles. \emph{Every Sample Matters}: a dual-track pipeline with Heterogeneous Triple-System Consensus evaluation and an Error Correction \& Derivation Module salvages every trajectory into usable supervision. \emph{Every Rollout Is Real}: a three-stage recipe---continual pre-training, supervised fine-tuning, and agentic reinforcement learning on hundreds of real phones---grounds every rollout in real production environments, so the capability the model learns transfers directly to deployment. \emph{Every Query Evolves}: a quota-driven benchmark methodology with three orthogonal axes enables precise attribution and allows the benchmark to be systematically upgraded as the model improves. \OurModel achieves \textbf{87.4} on \BenchPro, surpassing the best closed-source model by 5.1 points, and \textbf{84.9} on AndroidWorld, the best result among open-source models and competitive with closed-source models. These results demonstrate that grounding model training and iterative improvement in both real devices and the three Every principles yields strong, robust, and transferable mobile GUI capability.

\end{abstract}
\begin{document}

\maketitle

\begin{figure}[h]
    \centering
    \includegraphics[width=0.99\linewidth]{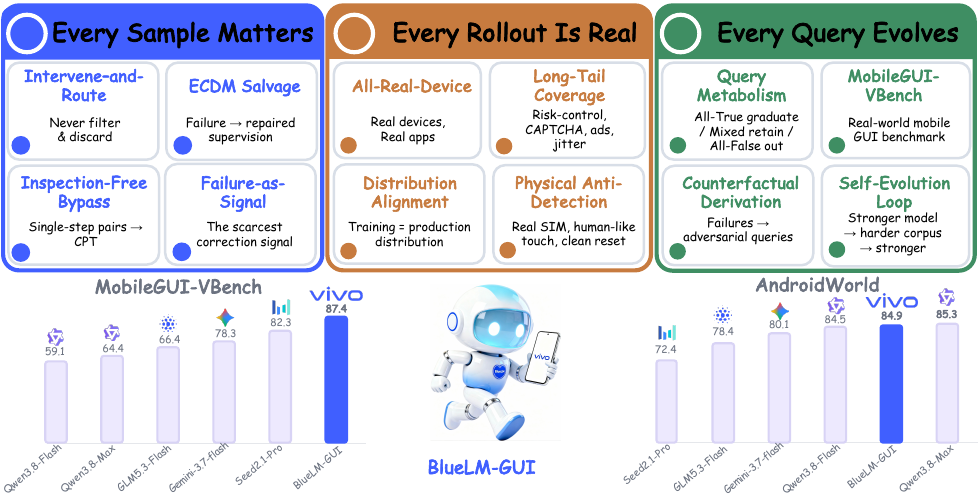}
    \caption{Overview of \OurModel, a real-device-centric flywheel built around three principles.}
    \label{fig:overview}
\end{figure}

\newpage

\vspace{0.5em}
{
  \hypersetup{linkcolor=black}
  \setlength{\parskip}{0pt}
  \renewcommand{\contentsname}{\normalfont\large\bfseries Contents}
  \setcounter{tocdepth}{3}
  \begingroup
    \small
    \tableofcontents
  \endgroup
}
\vspace{0.5em}

\newpage
\input{sections/introduction}

\input{sections/training_data}

\input{sections/model_training}

\input{sections/vivo_benchmark}

\input{sections/experiments}

\input{sections/related_works}

\input{sections/conclusion}

\input{sections/contributors}

\bibliography{references}

\newpage
\appendix
\input{sections/appendix.tex}

\end{document}

%% file: math_commands.tex
\usepackage{amsmath,amsfonts,bm}

\def\eqref#1{equation~\ref{#1}}

\def\1{\bm{1}}

\DeclareMathAlphabet{\mathsfit}{\encodingdefault}{\sfdefault}{m}{sl}
\SetMathAlphabet{\mathsfit}{bold}{\encodingdefault}{\sfdefault}{bx}{n}



%% file: sections/introduction.tex
\section{Introduction}
\label{sec:intro}

Recent advances in Large Language Models (LLMs)~\citep{deepseek-v4, gpt55, gemini31pro, claudeopus48} have approached or surpassed human-expert performance in natural language tasks~\citep{gpt55, kimik3} and coding~\citep{zhuo2025bigcodebench, terminalbench}, and across a wide range of downstream domains. This capability is rapidly extending along a curve from pure-text interaction toward vision-language-action multimodal agents. On this curve, mobile GUI agents occupy a uniquely critical position: mobile devices concentrate the largest and most frequently used entry points to digital services, where task-automation demand is explicit and commercial value is direct, making GUI agents one of the most industrially valuable directions among multimodal agents. Early systems were typically built as multi-module frameworks, composing planner, executor, and memory modules around general-purpose VLMs~\citep{mobileagent, appagent}; however, error propagation and information loss between modules cascade with task length, and multi-round module calls introduce unacceptable latency. Leading native GUI agents have therefore shifted to a different route: integrating perception, grounding, planning, and execution into a single model through post-training, allowing the model to learn decisions directly in an end-to-end closed loop~\citep{qwenuiagent, cao2026xiaomi, phonebuddy, hymobileagent}.

As the industry's GUI agent technology converges on the single-model paradigm~\citep{uitars, uitars2, Mobile-agent-v3, qwenuiagent}, the focus of system design shifts from module engineering to representation learning inside the model---and the latter depends largely on the quality of real agent data. This shift implies that what limits a mobile GUI agent today is no longer architectural sophistication, but problems that lie elsewhere: whether the training distribution matches the real distribution of production environments, and whether the benchmark can still accurately locate capability gaps after each iteration. The former concerns the data engine; the latter concerns the evaluation methodology. Yet in industrial practice, both remain fundamentally unsolved: the environments that generate training data are distributionally inconsistent with deployment~\citep{androidworld, androidlab, guiodyssey, mobilegym, simuwob}, the trajectories that carry the strongest signals remain under-utilized~\citep{singlerollout, uimopd, openclawrl}, and the benchmarks that steer iteration quietly saturate~\citep{androiddaily, mobileworld, phoneharness}. What truly differentiates leading systems is how fast they close---and keep turning---the loop of ``data production-training-evaluation-attribution-reproduction.''

Concretely, these problems manifest in practice as three gaps that stand in the way of closing this loop. The first is a distribution gap: the dominant approach trains in sandbox or emulator environments, or pairs them with real devices~\citep{androidworld, androidlab, guiodyssey, mobilegym, simuwob, maiui, hymobileagent, phonebuddy, cao2026xiaomi}. Sandboxes offer resettability and controllability, yet their environment composition, application states, and user-behavior distributions exhibit a substantial shift from real production environments and real usage. The second is a utilization gap: collecting trajectories on real devices is expensive and slow, and the resulting data contains a large share of failed or anomalous trajectories; these carry the scarcest correction and exploration signals, yet remain under-utilized~\citep{backtrackagent, mobile-r1}. The third is an attribution gap: under a self-evolution paradigm the model keeps improving, so any fixed benchmark saturates over time and loses its diagnostic power~\citep{claweval, workflowgym}---once a benchmark no longer separates good decisions from bad ones, its score stops telling us what to fix next. What is needed is not a better benchmark, but a \emph{method for producing benchmarks} that evolves together with the model. Motivated by these three gaps, we present \textbf{BlueLM-GUI}, a mobile GUI agent built as a real-device-centric flywheel for self-improving capabilities. We design data collection, model training, and evaluation as one closed loop governed by three complementary principles, one at each level of the system. We elaborate each principle in turn, following the pipeline they govern---data, training, and evaluation.

\paragraph{Every Sample Matters.}
Every collected trajectory is put to use---a direct response to the utilization gap above: failed and anomalous trajectories carry the rarest learning signal of all, so letting them slip away is a waste no system can afford. Successful trajectories enter the SFT pool as gold-standard supervision; failed or environmentally anomalous ones are routed to the Error Correction \& Derivation Module, which salvages them into new supervision through step-level correction, retroactive query alignment, or counterfactual query derivation. This principle governs the entire collection-and-training pipeline on our all-real-device farm: whether produced by guided collection or RL self-evolution, every trajectory is triaged by value---usable ones enter the pool, flawed ones are repaired and reborn, and those with no further incremental value retire naturally.

\paragraph{Every Rollout Is Real.} This is our answer to the distribution gap. For the training distribution to align with deployment, the most thorough way is to let training itself happen in the deployment environment---all of our RL exploration runs in live, interactive environments: every rollout is driven by real app executions and real system feedback. State transitions, interface responses, and anomalous events all come from the production distribution, so the capability the model acquires naturally fits deployment---a source of robustness that no simulator can offer. After each rollout, the trajectory flows back into the data pool: verifiable successes become supervision for later rounds, and failures tell us which tasks to sample next and thus the flywheel turns.

\paragraph{Every Query Evolves.} This is our answer to the attribution gap. Every query in our benchmark fills a quota-slot from an explicit metric system, rather than because an expert deemed it worth testing---so any score change is attributable to something concrete. Our benchmarks span three orthogonal axes---scenario, complexity, and interaction \& risk---with queries generated against quotas by an LLM and verified by experts. Trajectories are scored by multiple LLM judges that must rule unanimously; their disagreements go to human adjudication, whose verdicts refine judging rules. This design also makes benchmarks evolvable: when a benchmark saturates as \OurModel improves, we do not redesign a harder one---we shift quotas along path complexity, intent dependency, and instruction explicitness, yielding a new benchmark differing only in precisely stated quotas.

Empirically, the real-device flywheel translates into measurable gains. On our \BenchPro, \OurModel reaches \textbf{87.4}, surpassing the best closed-source model by 5.1 points and setting a new state of the art. On the emulator-based AndroidWorld benchmark~\citep{androidworld}, it scores \textbf{84.9}---the strongest among open-source and comparably sized models, and competitive with the largest closed-source model despite being orders of magnitude smaller. Together, these results show that grounding Every Sample, Every Rollout, and Every Query in real devices yields strong, robust, and transferable mobile GUI capability. Our contributions are threefold:

\begin{itemize}[leftmargin=1.6em]
\item \textbf{An all-real-device self-improving flywheel.} We present \OurModel, a mobile GUI agent whose entire training loop---RL exploration, rollout collection, and evaluation---runs on hundreds of real phones rather than sandboxes or emulators. Every rollout flows back into the data pool, closing a self-improvement loop whose training distribution coincides with deployment.

\item \textbf{A value-aware trajectory engine that turns failures into supervision.} We design an Error Correction \& Derivation Module that triages every trajectory by value: successes become gold-standard supervision, while failed and anomalous trajectories are salvaged into new supervision via step-level correction, retroactive query alignment, and counterfactual derivation. 

\item \textbf{A quota-driven benchmark methodology that evolves with the model.} Our \BenchPro is defined as a quota distribution over a metric system, enabling precise score attribution; when a benchmark saturates, shifting the quotas yields a new one whose difference is stated exactly.
\end{itemize}

The remainder of this report is organized as follows. \S\ref{sec:data} details the data engine, from query generation and real-device collection to trajectory evaluation, failure attribution, and the self-evolution loop. \S\ref{sec:training} presents the three-stage training recipe. \S\ref{sec:benchmark} describes the quota-driven benchmark methodology and its evaluation protocol. \S\ref{sec:exp} reports the main results and ablations that validate each principle, and \S\ref{sec:related} contrasts our approach with concurrent industrial work.

%% file: sections/training_data.tex
\section{Training Data Construction}
\label{sec:data}

\subsection{Data System Overview}
\label{sec:data-overview}

\begin{figure}[h]
    \centering
    \includegraphics[width=0.99\linewidth]{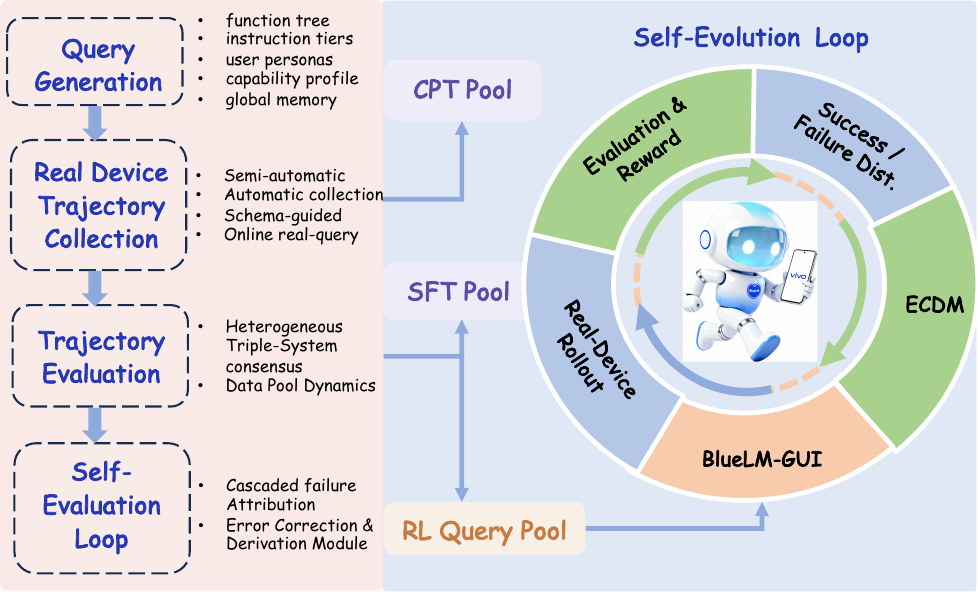}
    \caption{Overview of the BlueLM-GUI data system.}
    \label{fig:data-overview}
\end{figure}

\subsubsection{Design Philosophy}
\label{sec:data-philosophy}

BlueLM-GUI's data system is built on two complementary foundations: the first defines where the data comes from (the distribution), and the second defines how to make the most of every sample. The first is the all-real-device grounding that the training recipe (\S\ref{sec:training}) elevates into the principle \emph{Every Rollout Is Real}; the second is this section's own principle, \emph{Every Sample Matters}. Together, they make a real-device-only data pipeline not only feasible, but also capable of maximizing the data value.

\paragraph{Foundation 1: The Real-Device Distribution.}
All of our GUI trajectories come from real apps on real devices. The prevailing industrial approach pairs real devices with a large number of sandbox/mock environments, which offer resettability, instrumentation, and automated verification. However, they cannot reproduce the long-tail distribution that dominates production failures---risk-control challenges, CAPTCHAs, account states, popup ads, network jitter, and the version fragmentation of super-apps. Collecting on real devices aligns the training distribution with the production distribution and with real-world operating conditions---precisely the conditions under which mobile GUI agents fail in production. This choice is not free---all-real-device collection is time-consuming, hard to reset, and fragile to verify. This is precisely why Foundation~2 (``Every Sample Matters'') exists: to fully exploit every sample and maximize the value of real-device data.

\paragraph{Foundation 2: Every Sample Matters.}
Real-device collection is extremely expensive, so every trajectory and every evaluation result is expected to contribute value. Rather than the coarse filtering practice of ``keep the clean successes, discard the failures and anomalies,'' we extract value from every sample through an \textbf{intervene-and-route} philosophy. On the SFT side, trajectories evaluated as successful enter the SFT pool directly as gold-standard data, while trajectories judged as failed or suffering environmental anomalies are not discarded but routed to the \textbf{Error Correction \& Derivation Module} (ECDM, \S\ref{sec:data-ecdm}), which salvages them into new valid supervision. On the RL side, rollout results are likewise routed rather than filtered: boundary cases feed learning and even fully-failed rollouts flow to the ECDM for attribution and repair instead of being discarded (the pool-metabolism rule is detailed in \S\ref{sec:data-pool-dynamics}). The core idea is this: failures and anomalies are not waste; they are the scarcest correction and exploration signals.

\subsubsection{CPT/SFT Decoupling}
\label{sec:data-decoupling}

The two foundations above define where data comes from and how to make the most of every sample; a further question is how this scarce real-device data should be allocated across training stages. Our answer is to explicitly decouple the \textbf{CPT} (Continual Pre-Training) pool from the \textbf{SFT} (Supervised Fine-Tuning) pool (see Figure~\ref{fig:data-overview})---letting CPT absorb ``volume'' and SFT guard ``quality.''

\paragraph{Asymmetry 1: Production Speed.}
SFT data is inherently slow to produce: a trajectory must run end-to-end on a real device and pass the long acquisition--evaluation chain before it qualifies, so high-quality SFT data accumulates gradually. CPT data, by contrast, is produced at every collection step: single-step screenshot--action pairs enter the pool immediately after rule-based cleaning, without waiting for the final verdict, so a large volume accumulates quickly. If the two pools shared one intake gate, the fast stream would have to queue behind the slow stream and the model would starve for fresh data. Decoupling lets each stream run at its natural cadence: CPT continuously feeds the model with fresh single-step data at scale, while SFT quietly accumulates high-confidence gold-standard trajectories, and training can proceed with the former instead of stalling on the latter.

\paragraph{Asymmetry 2: Capability Foundation.}
Before the model can plan GUI tasks, it benefits from first acquiring a set of more general, low-level abilities: reading the screen (OCR, element localization), performing basic interactions, and calling tools (API invocation, parameter construction, result parsing, error handling). These abilities do not depend on long-horizon task success, so the CPT pool deliberately reserves a data ratio for two classes of foundational data: single-step screenshot--action pairs are valid even when the trajectory containing them ultimately fails, and tool-use data (API call pairs, tool schemas, parameter-construction examples) is injected to build a tool-calling prior during pre-training. With these general abilities in place after CPT, SFT can concentrate its scarce supervision on GUI planning itself---and on the decision of ``when to use a tool vs.\ when to operate the interface''---rather than teaching them from scratch. This also reserves the foundational capabilities for the subsequent GUI + API + CLI interaction paradigm.

\subsubsection{Dual-Track Pipeline}
\label{sec:data-dual-track}

The pipeline operates as two parallel tracks sharing the same real-device farm. The first track is the SFT data production line, organized as a four-ring main chain: query generation (\S\ref{sec:data-query-generation}), trajectory collection on real devices (\S\ref{sec:data-trajectory-collection}), trajectory-level evaluation via the Heterogeneous Triple-System Consensus (HTSC, \S\ref{sec:data-evaluation}), and data routing with self-healing. Verified trajectories enter the SFT pool, while failures flow to the ECDM for repair. The second track is the RL self-evolution line: queries of moderate difficulty are rolled out with the latest SFT model on real devices, and the results are routed by the within-group success-failure distribution (the pool-metabolism rule is detailed in \S\ref{sec:data-pool-dynamics}). The two tracks thus close a self-healing loop: SFT produces the model, RL probes its capability boundary, failures return through the ECDM as repaired supervision, and the cycle repeats.

The self-healing core spanning both tracks is the \textbf{Error Correction \& Derivation Module (ECDM)}. It attributes each failed or anomalous trajectory and converts it into new supervision through three repair paths---\textbf{step-level correction} (repair a single wrong step via a Teacher model with human fallback), \textbf{retroactive query alignment} (RQA, downgrade an over-scoped query to fit the correctly completed prefix), and \textbf{counterfactual derivation} (turn recurring failure classes into targeted adversarial queries for the RL pool). The full decision chain---attribution followed by the three paths---is described in \S\ref{sec:data-self-evolution}.

\subsection{Query Generation}
\label{sec:data-query-generation}

Query generation determines the starting point of the data pipeline and, to a large extent, the diversity of the trajectory samples---which in turn shapes the GUI model's generalization and robustness. Our design revolves around three dimensions: \textbf{functional coverage}, where the function tree systematically maps ``what to do'' to reduce blind spots; \textbf{instruction diversity}, where instruction tiers and user personas keep ``how it is said'' close to natural instruction distributions; and \textbf{task complexity}, where capability profiles together with Global Environment Memory elevate single-point tasks into complex tasks with genuine reasoning content.

\subsubsection{Function Tree}
\label{sec:data-function-tree}

Real commercial apps do not expose functionality as a flat list, but as a complex hierarchy of home-page entries, channel pages, search-result pages, detail pages, and action pages. We maintain a \textbf{function tree} for each client application, built through a semi-automatic pipeline:

\begin{itemize}[leftmargin=1.6em]
  \item \textbf{Automatic exploration}: an exploration agent automatically traverses the app, screenshotting each page layer and extracting its layout schema.
  \item \textbf{Semantic extraction}: a strong multimodal model (VLM) identifies UI interaction components, copy text, and latent functionality, automatically generating a JSON tree structure.
  \item \textbf{Human pruning}: humans prune conflicting and redundant nodes.
  \item \textbf{Reuse-relation tagging}: for the same functionality reachable via multiple paths (e.g., entering the same page via ``Me $\rightarrow$ Settings $\rightarrow$ Offline Packs'' and via ``Home $\rightarrow$ top-right corner $\rightarrow$ Offline Packs''), the tree marks a reuse relation to avoid duplicate expansion.
\end{itemize}

\begin{figure}[h]
    \centering
    \resizebox{0.85\linewidth}{!}{%
    \begin{forest}
      for tree={
        font=\footnotesize\ttfamily,
        grow'=0,
        child anchor=west,
        parent anchor=east,
        anchor=west,
        calign=first,
        edge path={
          \noexpand\path[\forestoption{edge}]
            (!u.parent anchor) -| ($(!u.parent anchor)!.5!(.child anchor)$) |- (.child anchor);
        },
        fit=rectangle,
        s sep=4pt,
        l sep=18pt,
      }
      [Translator Home, for tree={draw, rounded corners=2pt, fill=gray!10, minimum height=20pt, inner xsep=6pt, align=center}
        [Settings
          [Offline Translation Packs
            [English\\$\rightarrow$ Download, leafstyle]
            [Japanese\\$\rightarrow$ Download, leafstyle]
          ]
          [Announcement Settings, minimum width=141.94pt
            [Speech-speed\\slider, leafstyle]
          ]
        ]
        [Camera\\Translation, leafstyle]
        [Settings $\rightarrow$\\Offline Packs, reusestyle]
      ]
    \end{forest}%
    }
    \caption{An example function tree of a translator app. Green nodes are leaf targets; the dashed node marks a reuse relation.}
    \label{fig:function-tree}
\end{figure}
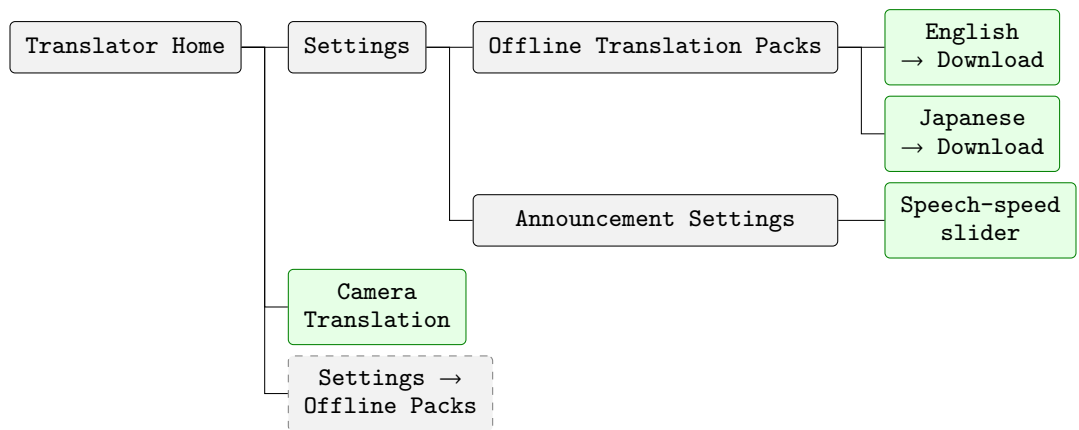
Figure~\ref{fig:function-tree} shows an example function tree for a translator app. The tree then serves as the unified index across the whole pipeline: query generation expands over leaf nodes, trajectory collection navigates toward them as targets, and coverage measurement computes the fraction of leaf nodes already exercised---the \textbf{function-tree coverage} metric, which turns ``full functional coverage'' into a measurable quantity.

\subsubsection{Instruction Tiers}
\label{sec:data-instruction-tiers}

To train the agent's adaptability to users with different levels of expression clarity, every leaf node on the tree is mapped to queries at three tiers. The higher the tier, the closer to real colloquial speech, and the more it tests the model's intent understanding and generalization.

\begin{itemize}[leftmargin=1.6em]
  \item \textbf{L1 step-by-step instruction}: the user knows the exact path and guides with explicit steps.
  \begin{itemize}[leftmargin=1.6em]
    \item \emph{Plain narration}: ``Tap Settings on the home page, tap Offline Translation Packs, and choose Download on the English card.''
    \item \emph{Colloquial steps} (same path, colloquial wording): ``First tap Settings on the home page, then pick that Offline Translation Packs thing inside, and finally download the English one.''
  \end{itemize}
  \item \textbf{L2 explicit goal-oriented instruction}: the user's goal is crystal clear, but no path is given. By semantic emphasis, we subdivide this tier into six subclasses for finer coverage:
  \begin{itemize}[leftmargin=1.6em]
    \item \emph{Function statement}: ``Go download an English offline translation pack.''
    \item \emph{Purpose statement} (implies the functional purpose): ``I'm about to board a plane and won't have internet---grab an English offline pack just in case.''
    \item \emph{Dependent operation} (long-chain before/after dependency): ``Download an English offline pack, and once it's done switch to real-time voice translation.''
    \item \emph{Parallel operation} (multi-tasking): ``Download an English offline pack, and while you're at it make the font on the translation screen a bit bigger.''
    \item \emph{Conditional operation} (branching): ``Check whether there's a Burmese offline translation pack in the app, and if there is, download it right away.''
    \item \emph{State/spatiotemporal-constraint operation}: ``I'm flying to Tokyo on a business trip the day after tomorrow---get me a usable offline pack for the local language.''
  \end{itemize}
  \item \textbf{L3 ambiguous / colloquial instruction}: simulates highly irregular colloquial expressions observed at high frequency.
  \begin{itemize}[leftmargin=1.6em]
    \item \emph{Incomplete instruction} (the goal is only hinted at): ``Get me that one that works without internet.''
    \item \emph{Intent inference} (no functional words at all): ``I'm heading abroad soon with no internet---how is this thing supposed to look up words?''
    \item \emph{Colloquial noise / self-correction} (the user changes the goal mid-sentence): ``Hey, that English one, download the local pack\ldots  wait no, I want the Japanese one!''
  \end{itemize}
\end{itemize}

\subsubsection{User Personas}
\label{sec:data-personas}

Even within the same instruction tier, different user groups exhibit distinctly different language habits. If the function tree determines what functionality is involved and the instruction tiers determine how the user says it, then the user personas determine who is saying it. We predefine eight standard user personas; during instruction generation, the LLM role-plays a specific persona and applies a style filter over the query's tone, wording, and habits for the same functionality point, suppressing the ``machine accent'' of synthesized data:

\begin{table}[h]
\centering
\small
\renewcommand{\arraystretch}{1.25}
\begin{tabular}{@{}m{0.22\linewidth}>{\centering\arraybackslash}m{0.22\linewidth}m{0.44\linewidth}@{}}
\toprule
\multicolumn{1}{c}{\textbf{Persona}} & \multicolumn{1}{c}{\textbf{Style}} & \multicolumn{1}{c}{\textbf{Example Query}} \\
\midrule
Average user / novice & Plain speech & ``Download the English offline pack---worried it won't work without internet.'' \\
College student / Gen Z & Internet slang & ``Quick, grab the offline translation---boarding soon and there's no signal!'' \\
Programmer & Terse and direct & ``English offline pack. Download.'' \\
Business professional & Formal and precise & ``Please download the English offline translation pack for use during my flight later.'' \\
Senior citizen & Rambling / anxious & ``Oh dear, the internet's about to cut out---will that translation thing still work? Hurry and save one for me.'' \\
Delivery rider & Urgent and clipped & ``English one, the offline kind. Quick.'' \\
Travel blogger & Enthusiastic and lively & ``Help! Flying to Thailand soon with no internet---save me a translation pack, quick!'' \\
Stay-at-home mom & Gentle, multitasking & ``Once the baby falls asleep, download the offline translation for me, and while you're at it check some baby-food recipes.'' \\
\bottomrule
\end{tabular}
\caption{The eight standard user personas and their example queries.}
\label{tab:data-personas}
\end{table}

\subsubsection{Capability Profile}
\label{sec:data-capability-profile}

The function tree determines which business domain is involved, and L1--L3 determine how the user says it (the expression axis). The \textbf{capability profile} adds the orthogonal complexity axis: it determines how hard the task is. Instruction tiers and user personas diversify expression without changing the intrinsic difficulty of the task, whereas the capability profile stacks reasoning and operational load onto the task, elevating a simple ``single-point functionality task'' into a ``multi-step, cross-app complex task demanding rigorous logical reasoning.''

To operationalize the complexity axis, we define six core capability dimensions: the first three raise the decision difficulty within a single point, and the last three stretch the task structure across pages and apps. When generating tasks, the LLM is guided to project tasks onto these dimensions:

\begin{itemize}[leftmargin=1.6em]
  \item \textbf{Multi-Constraint Filtering}: the model must identify and activate multiple filter conditions in the interface (e.g., in Meituan: ``find a Japanese restaurant with per-person spend under 100 CNY, a rating above 4.5, and delivery support'').
  \item \textbf{Numerical Reasoning \& Calculation}: the model must make mathematical decisions such as price comparison, currency conversion, and coupon stacking (e.g., ``select the first three items in the cart and figure out which combination with the spend-and-save coupon is the best deal'').
  \item \textbf{Conditional Logic}: the model chooses different execution paths based on runtime environment state (e.g., ``check whether this delivery order has free delivery---if yes, place the order; if not, switch to another restaurant'').
  \item \textbf{Cross-Page State Tracking}: the model must maintain short-term memory across long operation chains, using information extracted on Page A as input on Page C (e.g., ``remember the model number of that Sony headphone we just saw---we'll need it for price comparison later'').
  \item \textbf{Looping \& Iterative Execution}: the model must repeat actions over a series of similar components (e.g., ``download every PDF posted in the group chat today and save them all locally'').
  \item \textbf{Information Synthesis}: the model must gather scattered information across pages and apps and organize it into output (e.g., ``check Ctrip and Qunar separately for the cheapest flight to Beijing tomorrow, and send it to Zhang San on WeChat'').
\end{itemize}

The structural dimensions---cross-page state tracking, looping, and information synthesis---cannot be exercised within a single point or a single app, and even the decision dimensions compound in difficulty when combined across functionality points. The capability profile therefore pushes tasks toward cross-app, multi-functionality combinations---and this, in turn, imposes requirements on the physical reachability of synthesized instructions, which is exactly the problem addressed by Global Environment Memory in the next section.

\subsubsection{Global Environment Memory}
\label{sec:data-gem}

However, an LLM generating such cross-app tasks has no direct access to the real state of the target device. Synthesized cross-app instructions are therefore often physically unreachable: an LLM might fabricate ``read the tracking number in Xiaohongshu and paste it into Cainiao to query the parcel,'' yet Xiaohongshu's page may block screenshots due to risk control, or Cainiao may not be logged in on the current device at all. To bridge this gap, we maintain a \textbf{Global Environment Memory (GEM)}: a topological graph database that structures reachability knowledge on real devices, so that every step of a synthesized instruction has a physical foothold. GEM serves three roles in the pipeline: it is built from and updated by real-device exploration, it grounds cross-app query generation, and it receives failure feedback from the ECDM to keep itself fresh.

\paragraph{GEM Construction and Update Flow.}

GEM combines two complementary knowledge sources. The first is the page-navigation graph obtained from real-device exploration, built in four steps:

\begin{itemize}[leftmargin=1.6em]
  \item \textbf{Active environment exploration}: when the real-device farm is idle, exploration agents automatically perform non-destructive walkthroughs on real devices, recording page screenshots and component trees (Layout XML).
  \item \textbf{Perceptual deduplication}: pHash (perceptual hashing) + layout-similarity algorithms filter duplicate pages, keeping only page skeletons with unique functional states.
  \item \textbf{Navigation-graph construction}: build an adjacency navigation graph between pages. Edges come in two types:
  \begin{itemize}[leftmargin=1.6em]
    \item \emph{Intra-app navigation edges}: normal page navigation and interaction transitions.
    \item \emph{Cross-app navigation edges}: paths that switch via system-level media such as the Home key, the recents panel, the share sheet, or the clipboard.
  \end{itemize}
  \item \textbf{Functionality and data-flow compatibility extraction}: a strong VLM analyzes the ``input/output compatibility'' of each page (e.g., the output of a ``Taobao product-detail page'' is \code{Text:product title}; the input of a ``JD.com search page'' is \code{Text:search keyword}; the two are semantically type-compatible).
\end{itemize}

The second source is device- and system-level state knowledge, available to us as a device manufacturer: functionality differences across device models and OS versions, exact paths to system settings, and the capability boundaries of pre-installed apps. This knowledge is injected into GEM as precise priors on functionality existence and paths, complementing the verified navigation reachability from real-device exploration. Together, the two sources ensure that synthesized queries are tailored to the target device's real environment, rather than exposing path mismatches only at collection time.

\paragraph{Cross-App Task Generation Flow.}

With GEM, generating complex cross-app instructions such as ``price comparison'' and ``relay'' tasks shifts from ungrounded guessing to grounded generation after graph-path retrieval:

\begin{itemize}[leftmargin=1.6em]
  \item \textbf{Choose a start point}: randomly sample a start point from the function tree (e.g., search headphones on Taobao).
  \item \textbf{Path exploration}: query the GEM graph database for a cross-app endpoint whose input/output is compatible with the start point (e.g., JD.com search).
  \item \textbf{Path-existence check}: verify that the physical path ``Taobao $\rightarrow$ launcher $\rightarrow$ JD.com'' exists and is traversable.
  \item \textbf{Generate grounded query}: the VLM combines the real page contexts of both apps to generate a reachable query.
\end{itemize}

\paragraph{Closed-Loop Interaction with the ECDM.}

When a real device hits an \code{Env\_ERROR} while executing an instruction, the failure signal flows back to GEM: the broken path is marked ``temporarily unreachable'' and instruction generation over that path is halted, while query generation reroutes to a reachable alternative path in the updated graph. This keeps GEM synchronized with the ever-changing real-device environment.

\subsubsection{Query Intake Processing}
\label{sec:data-query-intake}

Queries reach the collection stage through two channels---newly synthesized and externally imported---and each channel receives a different treatment beforehand: synthesized queries pass a quality gate at generation time, while external stock queries are post-hoc tagged to fit the same label system.

\paragraph{Quality Gate for Newly Synthesized Queries.}

\begin{itemize}[leftmargin=1.6em]
  \item \textbf{Gate 1: Three-axis evaluation by an LLM judge}. Newly generated queries must pass an independent LLM judge on three dimensions:
  \begin{itemize}[leftmargin=1.6em]
    \item \emph{Realism}: whether the query conforms to genuine human expression habits (hard filter).
    \item \emph{Completability}: whether it is physically executable given the functionality currently present in mainstream apps (hard filter; failures are discarded outright).
    \item \emph{Complexity}: evaluates operation-step count and reasoning depth; not used as a hard filter, but stored as a metadata label for fine-grained mixing ratios and RL-pool candidate screening (\S\ref{sec:data-pool-dynamics}).
  \end{itemize}
  \item \textbf{Gate 2: Natural-language polishing (denoising \& polishing)}. A lightweight rewriting model converts text with a heavy ``LLM-prompt accent'' into idiomatic expression with natural pauses, colloquial noise, and abbreviations (echoing the persona styles of \S\ref{sec:data-personas}).
  \item \textbf{Gate 3: Back-tagging to function-tree nodes}. An algorithm automatically anchors each approved query back to the corresponding leaf or subtree node of the function tree (\S\ref{sec:data-function-tree}), enabling real-time monitoring of function-tree coverage and node-weighted training mixtures.
\end{itemize}

\paragraph{Post-hoc Four-Dimensional Tagging of External Stock Queries.}
For stock queries from curated external sources, business BadCases, or open-source datasets---which were not produced by this pipeline and carry no production labels---we run a round of four-dimensional post-hoc tagging with an LLM, so that they become comparable and mixable with internally produced queries under one unified label system:

\begin{itemize}[leftmargin=1.6em]
  \item \textbf{Step depth}: the estimated minimum number of operation steps required.
  \item \textbf{Ambiguity grade}: aligned and classified into L1 / L2 / L3.
  \item \textbf{Risk grade}: flags whether the query involves high-risk actions such as financial payment, privacy exposure, or sensitive account-information modification.
  \item \textbf{Pre-requisites}: whether the query depends on specific physical preconditions (e.g., ``bank card bound,'' ``account logged in''), so the real-device farm can automatically preset the environment before execution.
\end{itemize}

\subsection{All-Real-Device Trajectory Collection}
\label{sec:data-trajectory-collection}

After query generation, the next step is to turn an abstract query into a high-quality real-device trajectory carrying a fine-grained reasoning chain. We built an all-real-device farm hosting three parallel collection lines plus one auxiliary external query channel (\S\ref{sec:data-collection-methods}); every trajectory carry structured reasoning chains (\S\ref{sec:data-structured-cot}) and then pass horizontal enrichment (\S\ref{sec:data-enrichment}) before entering the pools.

\subsubsection{Trajectory Collection Methods Overview}
\label{sec:data-collection-methods}

\begin{figure}[h]
    \centering
    \includegraphics[width=0.98\linewidth]{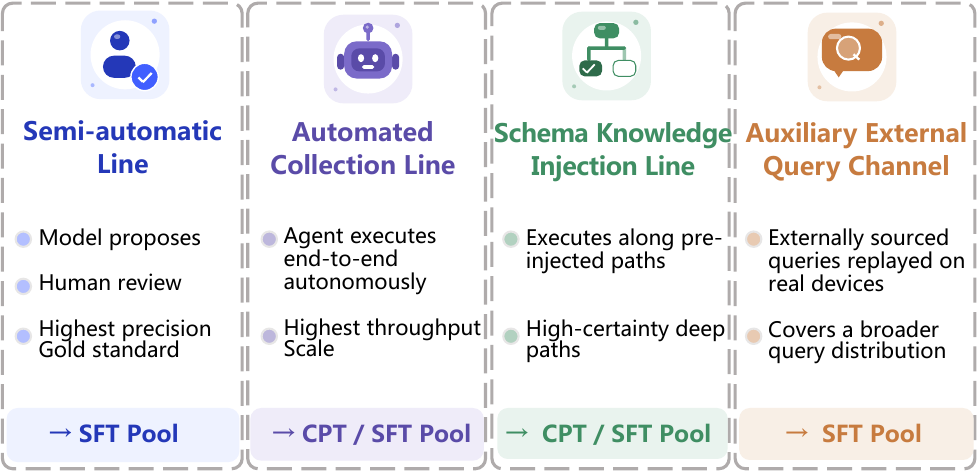}
    \caption{The four collection channels of the BlueLM-GUI data system: three parallel collection lines (semi-automatic, automatic, schema-guided) plus one auxiliary external query channel.}
    \label{fig:data-four-line}
\end{figure}

Different tasks demand different trade-offs between precision, cost, and throughput, so we run four collection channels on the all-real-device farm---three parallel lines plus one auxiliary external query channel---as illustrated in Figure~\ref{fig:data-four-line}:

\begin{itemize}[leftmargin=1.6em]
  \item \textbf{Semi-automatic line} (model proposes, human verifies each step): highest precision, for gold-standard data---verified trajectories go straight to the SFT pool (\S\ref{sec:data-semi-automatic}).
  \item \textbf{Automatic line} (agent policy runs end-to-end, unsupervised): highest throughput, for scale---raw trajectories are routed by HTSC evaluation (\S\ref{sec:data-automatic}).
  \item \textbf{Schema-guided line} (model follows a pre-injected operation path): for high-determinism deep flows (system settings, vertical apps) where blind exploration collapses---trajectories are routed by HTSC evaluation (\S\ref{sec:data-schema-guided}).
  \item \textbf{Auxiliary external query channel} (externally sourced queries replayed on real devices): for covering a broader query distribution---replayed trajectories are routed by HTSC evaluation (\S\ref{sec:data-online-channel}).
\end{itemize}

\subsubsection{Semi-Automatic Collection}
\label{sec:data-semi-automatic}

Semi-automatic collection is our primary gold-standard production line. It adopts a ``model proposes, human verifies (HITL)'' collaboration mechanism: the model contributes fast reasoning, and the human ensures step-level correctness. Because a human performs ``surgical'' verification at every step, this line keeps hallucinated trajectories out of the data and preserves the device time invested in the preceding successful steps.

\begin{itemize}[leftmargin=1.8em]
  \item \textbf{Dual-mode recommendation}: the annotation client pulls the current real-device screenshot and queries the LLM, which outputs a structured Chain-of-Thought (CoT) and a predicted action.
  \item \textbf{Micro-adjustment verification}: if the recommended action is accurate, the verifier confirms with one click; if it deviates (e.g., the tap position is off by tens of pixels, or a ``slider drag'' is misjudged as a ``tap''), the verifier directly drags the tap box to fine-tune it, modifies the action type, or rewrites the text on the screen image to correct it.
  \item \textbf{Execution and iteration}: the system executes the verified action on the real device, captures the new screen, updates the state, and proceeds to the next step until the task completes.
\end{itemize}

\subsubsection{Automatic Collection}
\label{sec:data-automatic}

The automatic line trades per-step verification for throughput: the agent model drives tasks end to end, so collection volume is far higher than the semi-automatic line, at the cost of step-level correctness, which physical-level safeguards and post-hoc replay correction compensate for.

\begin{itemize}[leftmargin=1.6em]
  \item \textbf{Closed-loop execution flow}: the scheduling hub dispatches a query, and the model autonomously runs the observe--decide--act loop, judging for itself whether the task is complete or ending on a stop condition (timeout, maximum steps exhausted, or consecutive repeated actions).
  \item \textbf{Physical-level self-cleaning and safety isolation}:
  \begin{itemize}[leftmargin=1.6em]
    \item \textbf{Physical state reset}: before each task, the control system force-stops all background processes (\code{am force-stop}) and clears the target app's data (\code{pm clear}) via low-level ADB commands, so every launch starts from a clean initial state.
    \item \textbf{Physical anti-risk-control disguise}: devices are bound to dedicated physical SIM cards on independent cellular networks; instead of ``teleport''-style coordinate taps with no transition trajectory, the system injects touch-trajectory signals with realistic physical acceleration, evading commercial apps' device risk-control engines.
  \end{itemize}
  \item \textbf{Post-hoc triage and repair routing}: without per-step verification, automatic trajectories may mix in cases of mid-task disorientation, misjudged completion, and interruptions by popups or ads. These trajectories are flagged and funneled into the HTSC evaluation routing of \S\ref{sec:data-evaluation}.
\end{itemize}

\subsubsection{Schema-Guided Collection}
\label{sec:data-schema-guided}

On-device settings and deep vertical apps (e.g., banking apps, developer options) have deterministic yet complex operation paths, where slight visual or copywriting differences can collapse a blind exploration's success rate. We therefore developed the \textbf{Schema-Guided Rollout} line: the known operation path of a functionality point is injected into the system prompt before execution, so the model follows the path instead of exploring. It supports two modes:

\begin{itemize}[leftmargin=1.6em]
  \item \textbf{Tap-operation mode}: inject the complete operation path (e.g., ``Settings $\rightarrow$ WLAN $\rightarrow$ Advanced Settings $\rightarrow$ MAC Address''); the model taps through it step by step, skipping trial-and-error exploration; when it encounters an irreversible toggle-type operation or a system confirmation dialog, it emits a clarification to the user and proceeds only after confirmation.
  \item \textbf{Search-operation mode}: inject a search term plus a functional-location description (e.g., ``search `Bluetooth' in Settings, tap the second result''); the model locates the target page through the search box instead of drilling down layer by layer.
\end{itemize}

\subsubsection{Auxiliary External Query Channel}
\label{sec:data-online-channel}

Beyond the three synthesized-query lines, we operate an additional channel that brings in externally sourced queries, complementing the synthesized-query distribution:

\begin{itemize}[leftmargin=1.8em]
  \item \textbf{Query pull}: pull queries for GUI-skill tasks from curated external channels.
  \item \textbf{Cleaning and intent classification}: all queries undergo multi-stage anonymization and privacy audit, followed by LLM-based cleaning and multi-level intent classification, removing invalid, sensitive, and non-executable samples.
  \item \textbf{Real-device automatic replay}: replay these queries on real devices via automatic execution tools, producing structured trajectories with step-by-step \code{cot / action}.
  \item \textbf{Result routing}: replayed trajectories enter the HTSC evaluation of \S\ref{sec:data-evaluation} just like synthesized-line trajectories---those that pass are corrected and enter the training pool, while failures go to the ECDM for attribution and repair.
\end{itemize}

\subsubsection{Structured CoT}
\label{sec:data-structured-cot}

GUI task execution involves long-horizon sequential decisions; a model trained purely on action sequences cannot generalize its reasoning to previously unseen pages. We therefore require every collected trajectory to carry a structured reasoning chain before each action, embedding both ``what to do (What)'' and ``why (Why)'' directly in the data. The schema is written identically into the system prompt at inference time, making it the data foundation for process supervision in the subsequent SFT phase.

\paragraph{Five-Tag Structured CoT.}

Each action produced during collection is preceded by the five tags (Table~\ref{tab:data-structured-cot}):

\begin{table}[h]
\centering
\small
\renewcommand{\arraystretch}{1.25}
\begin{tabular}{@{}>{\centering\arraybackslash}m{0.14\linewidth}m{0.34\linewidth}m{0.44\linewidth}@{}}
\toprule
\multicolumn{1}{c}{\textbf{Tag}} & \multicolumn{1}{c}{\textbf{Meaning}} & \multicolumn{1}{c}{\textbf{Example}} \\
\midrule
\code{[Observation]} & Description of the current page state & ``Currently on the Taobao search-results page; a search box at the top, a product list below'' \\
\code{[Reflection]} & Review of and correction to the previous step (triggered only when the observation conflicts with the previous step's plan) & ``Tapped the filter button last step, but the page didn't change---the tap position may have been off'' \\
\code{[Plan]} & Next-step plan & ``Tap the search box and type `Sony XM5' to search'' \\
\code{[Decision]} & The concrete action & ``Tap the search box at coordinates (540, 200)'' \\
\code{[Memory]} & Information to remember across steps & ``Remember: the product currently being compared is the Sony XM5, priced at 1,999 CNY'' \\
\bottomrule
\end{tabular}
\caption{The five tags of the structured CoT.}
\label{tab:data-structured-cot}
\end{table}

\paragraph{Bi-Directional Consistency Constraint.}

To prevent ``post-hoc rationalization'' (generating the action first, then assembling plausible-looking reasoning around it), a bi-directional verification gate runs before trajectories are replayed into the pool:

\begin{itemize}[leftmargin=1.6em]
  \item \textbf{Forward derivability}: given \code{[Observation]} + \code{[Plan]}, the \code{[Decision]} must be the uniquely reasonable action---different actions must not all ``make sense.''
  \item \textbf{Label consistency}: the action in \code{[Decision]} must exactly match the action executed in the trajectory; the page state in \code{[Observation]} must match the screenshot; the \code{[Memory]} items must actually be referenced in later steps. Any violation triggers regeneration until all pass.
\end{itemize}

\subsubsection{Horizontal Enrichment}
\label{sec:data-enrichment}

Raw trajectories collected from each line pass through a unified horizontal enrichment pipeline, which reorganizes and refines existing trajectories or intervenes during collection without creating new tasks, so that the model gains stronger generalization and robustness. The pipeline comprises three classes of processing:

\begin{itemize}[leftmargin=1.6em]
  \item \textbf{Step-level context disambiguation}: different apps can have extremely similar visual structures (e.g., the near-identical ``shopping cart'' pages of various supermarket apps). In each step's trajectory metadata, we explicitly write the current system-level state obtained via ADB at collection time---the foreground app package name (e.g., \code{com.taobao.taobao}) and the Activity stack---giving the VLM an unambiguous physical identifier for multi-dimensional reasoning in cross-app scenarios and reducing the drift of purely visual reasoning.
  \item \textbf{Anomalous state injection}: to train an agent that keeps functioning under harsh conditions, we inject two classes of unexpected disturbances during real-device collection---\emph{system interruptions} (mid-operation push notifications, low-battery alerts, or pre-set permission dialogs such as ``Allow location access'') and \emph{network fluctuations} (artificially induced latency or disconnection that forces a ``Network unavailable, please retry'' placeholder). Capturing the model's successful self-recovery paths under these disturbances (e.g., tapping ``Allow'' on a dialog, or tapping ``Refresh'') builds its defensive playbook. For anomalous states that cannot be injected---login-session expiry, CAPTCHAs, payment verification---we collected roughly 5{,}000 targeted samples covering 14 classes: the model learns to stop and hand back control where human intervention is required (e.g., CAPTCHAs), and to close or bypass safely skippable pages (ads, non-critical popups).
  \item \textbf{Repetition-to-reflection}: real-device rollout trajectories often contain repeated segments where ``a single action or a small group of actions is executed multiple times in a row'' (from local failure modes or page lag / network jitter). Used directly as training labels, these would teach the model to ``repeat when stuck.'' We remove the repeated steps from the training labels but keep them in the trajectory history (as evidence of ``possible stalling''); when a later step breaks out of the stall (going back / switching entry / replanning), that corrective step remains a label, with the earlier failed attempts still in its context---turning a harmful label pattern into a reflection- and error-recovery-oriented supervision signal that works in concert with the \code{[Reflection]} field of the structured CoT (\S\ref{sec:data-structured-cot}).
\end{itemize}

\subsection{Trajectory Evaluation and Data Routing}
\label{sec:data-evaluation}

In an industrial-grade GUI-agent data pipeline, trajectory-quality judgment and efficient data routing form the hub that largely determines how far model capability can evolve. We therefore designed a \textbf{two-phase routing architecture} paired with \textbf{Heterogeneous Triple-System Consensus (HTSC) evaluation}, ensuring that every trajectory---whether collected on real devices or rolled out autonomously---is precisely routed and fully utilized, putting the ``Every Sample Matters'' philosophy into practice.

\subsubsection{Two-Phase Evolution of Data Routing}
\label{sec:data-two-phase}

A query's routing difficulty can only be measured against an existing policy (e.g., ``moderate difficulty'' defined by success rate), yet no such policy exists at cold start. We therefore split the routing mechanism into two phases according to whether a current policy exists. The HTSC verdict criterion itself is phase-invariant---it always judges whether the trajectory physically completed the task; what evolves across phases is the difficulty yardstick for routing, which applies to the RL query pool:

\begin{itemize}[leftmargin=1.6em]
  \item \textbf{Phase 0 $\cdot$ Bootstrap}: no usable GUI policy model yet. Trajectory routing relies on policy-independent objective assertions only: HTSC judges correctness, and the three data pools fill according to their fixed rules---the CPT pool accumulates single-step screenshot--action and tool-use corpus via its inspection-free bypass (not routed by HTSC; \S\ref{sec:data-pool-dynamics}), while the SFT pool accumulates gold-standard corpus from HTSC-verified and human-adjudicated trajectories, and the RL query pool is pre-filled by objective difficulty features (its phase-by-phase metabolism is detailed in \S\ref{sec:data-pool-dynamics}). The core goal of this phase is to train the initial policy model $M^{(0)}$.
  \item \textbf{Phase 1 $\cdot$ Self-Evolution}: the current policy model $M^{(t)}$ is now in service. The difficulty yardstick upgrades to policy-dependent: the actual rollout success/failure distribution (success rate) of $M^{(t)}$ over the query pool serves as the yardstick, selecting samples with the highest learning signal at the current capability boundary to drive the coordinated iteration of RL and SFT (see \S\ref{sec:data-pool-dynamics}).
\end{itemize}

\subsubsection{Heterogeneous Triple-System Consensus}
\label{sec:data-htsc}

In the overall flow diagram (Figure~\ref{fig:data-overview}), ``trajectory evaluation'' is the pivotal checkpoint. A single evaluator---however capable---suffers from multimodal hallucination and single-perspective blind spots, so its verdict cannot be trusted as the sole gatekeeper of training-pool purity. We therefore built the Heterogeneous Triple-System Consensus (HTSC) mechanism: three evaluators with heterogeneous technology stacks cross-validate the same trajectory under a shared safety veto, a consensus decision maker aggregates their verdicts into one of four outcomes, and each outcome maps to a fixed routing destination. This section describes these stages and the downstream routing in order.

\paragraph{Stage 1 $\cdot$ Safety Veto (pre-screen).}

Safety is enforced by a shared safety checklist that all three systems execute in parallel with their correctness evaluation---covering scenarios such as direct payment, sensitive permissions, privacy-data access, and irreversible dangerous operations. A safety violation detected by any system triggers a one-vote veto: the trajectory bypasses the consensus vote entirely, is judged \code{False} outright, and is routed to the ECDM---safety red lines do not participate in majority voting, and there is no channel through which ``most systems see no problem'' can override.

\paragraph{Stage 2 $\cdot$ Three Heterogeneous Evaluators.}

The three systems divide the work along different evaluation dimensions, cross-validating the same trajectory from three levels---terminal state, key milestones, and overall path---and reduce the risk of common-mode failure through heterogeneity of the underlying technology stacks:

\begin{itemize}[leftmargin=1.6em]
  \item \textbf{System A (terminal-state physical verification)}: focuses on factual verification via system-level data---reading OS-level APIs, database logs, network traffic, and filesystem state---to make hard-metric assertions about the task's terminal state (e.g., ``has the file been downloaded,'' ``has the setting been enabled''). Highest decision certainty, but only covers tasks whose terminal state is readable from the device side.
  \item \textbf{System B (milestone-decomposition verification)}: focuses on two-granularity ``milestone + global'' analysis---first decomposing the task into an ordered subgoal checklist, checking along the trajectory whether each milestone is reached, then making a global judgment on the overall result. It answers ``were all the key nodes traversed correctly.''
  \item \textbf{System C (overall-path plausibility verification)}: focuses on semantic-level evaluation of the overall task and planned path---consuming the long-horizon Action-Observation history and judging whether the trajectory's progression logic is coherent, and whether there is meaningless detouring or pseudo-completion (appearing to progress while not actually reaching the goal). It answers ``is this path reasonable.''
\end{itemize}

\paragraph{Stage 3 $\cdot$ Consensus Voting.}

The consensus decision maker aggregates the three systems' verdicts into one of four outcomes---\code{True} (task completed), \code{False} (model execution failure), \code{Env\_ERROR} (failure attributable to environmental factors), or \code{Undetermined} (no consensus). Each system votes from within the label space \{\code{True}, \code{False}, \code{Env\_ERROR}, \code{abstain}\} (System A abstains when it cannot read the terminal state; \code{Env\_ERROR} is flagged upon detecting popup interception, network anomalies, missing permissions, or app crashes). Voting proceeds by the following priority:

\begin{itemize}[leftmargin=1.8em]
  \item \textbf{System A can read the terminal state}: A's physical assertion dominates the decision as a high-confidence fact. An A assertion of failure yields \code{False} (or \code{Env\_ERROR} if A identifies an environmental cause) regardless of B and C; an A assertion of success yields \code{True} unless B and C unanimously object, in which case the outcome is \code{Undetermined}.
  \item \textbf{System A cannot read the terminal state (abstains)}: Systems B and C vote---unanimous pass yields \code{True}, unanimous failure yields \code{False}, and disagreement yields \code{Undetermined}.
\end{itemize}

\paragraph{Routing of the Four Outcomes.}

Each consensus outcome maps to a fixed destination---the verdict decides where the trajectory goes, and no trajectory is ever discarded:

\begin{itemize}[leftmargin=1.8em]
  \item \textbf{True $\rightarrow$ SFT pool}: a high-quality successful trajectory, injected directly into the SFT pool.
  \item \textbf{False $\rightarrow$ ECDM}: confirmed algorithmic execution failure, sent to the correction module for step-level repair or retroactive alignment (\S\ref{sec:data-ecdm}), after which it becomes high-quality SFT data.
  \item \textbf{\code{Env\_ERROR} $\rightarrow$ ECDM}: an objective failure caused by environmental factors---not counted as a model error. It triggers the environment self-healing mechanism and re-initiates collection; meanwhile, the correction module can synthetically derive ``close the popup and resume the task'' repair steps into the trajectory, recovering usable supervision from these samples.
  \item \textbf{Undetermined $\rightarrow$ human adjudication}: detailed below.
\end{itemize}

\paragraph{Human Adjudication of Undetermined Samples.}

An \code{Undetermined} sample should neither rashly enter the gold-standard pool and pollute SFT, nor rashly be sent to correction as a failure, so it is routed to human adjudication: human judges correct $\rightarrow$ SFT pool; human judges wrong $\rightarrow$ ECDM (\S\ref{sec:data-ecdm}); human judges environment issue $\rightarrow$ the \code{Env\_ERROR} branch with environment self-healing and re-collection. Beyond settling individual samples, this loop compounds: disagreement samples are the evaluation system's most valuable ``stress test,'' concentrating exposure of single-modality judges' blind spots, and the human verdicts are simultaneously used to retroactively calibrate the thresholds and prompts of the evaluation systems, so HTSC's consensus rate keeps rising across iterations. This routing design puts Every Sample Matters into practice: a verdict disagreement is itself a supervision signal, not noise to be smoothed away by majority vote.

\subsubsection{Data Pool Dynamics}
\label{sec:data-pool-dynamics}

Evaluation (above) together with attribution and ECDM repair (\S\ref{sec:data-self-evolution}) yield finished data; this section answers its final destination: the three data pools. What separates them is less \emph{what} they store than whether a sample's value persists or shifts as the model $M^{(t)}$ evolves---two pools accumulate monotonically, while the RL pool must metabolize.

\begin{itemize}[leftmargin=1.6em]
  \item \textbf{CPT pool---static accumulation, append-only}. The single-step ``pixel--element--action'' correspondence is independent of the trajectory's final success (\S\ref{sec:data-decoupling}), and its correctness does not change with the policy either: a valid single-step sample is equally valid for $M^{(0)}$ and for $M^{(t)}$. The pool therefore only appends, never retires---massive single-step data flows in inspection-free and accumulates continuously.
  \item \textbf{SFT pool---static accumulation, purity first}. It receives only three classes of high-confidence trajectories (HTSC-verified \code{True}, human-adjudicated correct, and ECDM-repaired qualified; \S\ref{sec:data-htsc}, \S\ref{sec:data-ecdm}). Correctness, once confirmed, holds indefinitely, so this pool likewise grows monotonically; it differs from the CPT pool only in intake-gate strictness, not in temporal behavior.
  \item \textbf{RL query pool---dynamic metabolism, difficulty moves with the model}. The one exception: a query's RL value depends not on the query itself, but on whether the current model can solve it. As the model strengthens, yesterday's boundary case becomes today's easy win, and the pool must metabolize accordingly---otherwise RL compute is wasted on already-mastered queries.
\end{itemize}

The RL pool's metabolism therefore proceeds in two phases (motivation in \S\ref{sec:data-two-phase}):

\begin{itemize}[leftmargin=1.6em]
  \item \textbf{Phase 0 (Bootstrap)}: no policy exists to serve as a difficulty yardstick, so policy-independent objective features stand in---instruction tier L2/L3, large step depth, cross-app linkage, multi-functionality combinations. Naturally difficult queries are pre-screened and banked in advance, to be activated once $M^{(0)}$ is trained.
  \item \textbf{Phase 1 (Self-Evolution)}: after each round of RL training, $M^{(t)}$ performs multiple rollouts over the pool's queries, and the within-group success/failure distribution drives a three-way metabolism at the query level: All True $\rightarrow$ retirement (mastered, no learning signal remains); mixed outcomes $\rightarrow$ retention (at the capability boundary, highest gradient information gain); All False $\rightarrow$ the query leaves the RL pool and its failed trajectories flow to the ECDM (beyond the boundary; repaired into SFT corpus to help broaden it).
\end{itemize}

Intake into the RL pool comes from multiple channels: cold-start objective-feature screening, data-pipeline routing, business-side supplementation, and targeted supplementation of BadCases exposed online / in evaluation. Under Phase 1, all channels pass a uniform gate---``sort by the current policy's success rate and remove all-true / all-false''---ensuring every entering query falls in the medium-success-rate zone of the current policy.

It is precisely this structure---two static pools as the foundation, one dynamic pool probing the boundary---that lets the data system both accumulate (correctness endures) and keep pace (difficulty moves with the model): the reservoir on which the self-evolution loop of the next section runs.

\subsection{The Self-Evolution Loop}
\label{sec:data-self-evolution}

The preceding sections (\S\ref{sec:data-query-generation}--\S\ref{sec:data-evaluation}) are ``one-shot'' data production: queries are generated, trajectories collected, evaluated, and routed into pools---producing a batch of static corpus. But a GUI agent's capability ceiling depends on whether it can continuously update its data as it improves. The self-evolution loop introduced in this section is the engine of ``Phase 1'': starting from the initial model $M^{(0)}$ trained during Phase-0 bootstrap, it drives the model and the corpus to co-evolve across rounds.

\paragraph{Limitations of Static Corpus.}

GUI execution is sequential decision-making: a single mis-tap at step $i$ changes the state distribution at step $i+1$, and the error cascades through all subsequent steps. This gives static corpus two structural limitations: supervision is mostly \textbf{off-policy} (the state distribution of annotated demonstrations mismatches the distribution the current model actually visits) and \textbf{success-biased} (``correct actions in correct states'' are over-represented). The model thus only learns ``how to continue on the right track,'' while supervision is thin on ``how to recognize a wrong state'' and ``how to recover''---the main bottleneck of reflection ability and complex-task performance.

The loop addresses the two limitations to different degrees. \textbf{Success bias} is covered directly: rollout failures are no longer discarded but converted by the ECDM into three classes of supervision---single-step correction labels within erroneous states, short-chain demonstrations anchored on correct prefixes (RQA), and adversarial tasks targeting weak capabilities---so failure and recovery signals continuously enter the training distribution. \textbf{The off-policy gap} is covered progressively: ECDM-repaired labels remain offline rewrites, but each round starts from rollouts of the current model $M^{(t)}$, so the repaired supervision is anchored on the states the current policy actually visits, and the train--visit distribution gap narrows across rounds; the RL line further provides strictly on-policy gradients, complementing the SFT line.

\paragraph{Single-Loop Self-Evolution Dynamics.}

Once $M^{(0)}$ is trained in Phase 0, the loop runs at full capacity, each iteration executing four serial steps:

\begin{center}
\textbf{Model rollout $\rightarrow$ evaluation and cascaded attribution $\rightarrow$ error correction \& synthetic derivation (ECDM) $\rightarrow$ retrain and re-rollout}
\end{center}

The loop is error-driven: rollout blind spots (All False and \code{Env\_ERROR}) become ECDM input, and the repaired corpus upgrades the model to $M^{(t+1)}$, which attacks harder boundary cases---model and corpus co-evolve across rounds. Among the three destinations of the attribution step---environment self-healing, evaluation-system revision, and ECDM repair---it is the third that closes this loop (the right half of Figure~\ref{fig:data-overview}). As iterations proceed, simple errors disappear from rollouts and correction data tracks the model's capability ceiling. We now detail the loop's two core steps in turn: cascaded attribution (\S\ref{sec:data-attribution}), which decides where each failure goes, and the ECDM (\S\ref{sec:data-ecdm}), which turns it into data.

\subsubsection{Cascaded Failure Attribution}
\label{sec:data-attribution}

When a trajectory flows into the \code{False} or \code{Env\_ERROR} branch, the first step of the ECDM is cascaded attribution: failure causes are determined level by level, in a fixed order---engineering side first, algorithm side last---and the conclusion at each level directly decides the trajectory's subsequent processing path within the ECDM. The cascade is necessary because environmental interference and model deficiencies are often isomorphic in their trajectory-level appearance: a mid-task interruption caused by a network disconnection and one caused by lost state tracking are nearly indistinguishable in the Action-Observation history. Without first excluding objective environmental factors, environment noise would be misattributed to capability gaps, and the ECDM would then generate repair directions aimed at the wrong target.

The attribution engine executes three levels of diagnosis in the following order:

\begin{itemize}[leftmargin=1.8em]
  \item \textbf{Level 1: environment-failure check (objective, device-side facts)}. Examine whether the trajectory contains observable environmental interference---system-popup interception, network anomalies, missing permissions, app crashes. If the failure is directly attributable to such factors, it is labeled \code{Env\_ERROR}: the trajectory is not counted as a model error, and the structured interference features (popup type, occurrence timing, network state) are extracted and fed back to the environment-maintenance module, triggering automated self-healing of the real-device environment and re-collection.
  \item \textbf{Level 2: task- and verifier-defect check (data-production-side)}. With environment factors excluded, examine whether the failure originates from the data-production side rather than the model, separating two sub-cases by owner: a \emph{task-specification defect}---an ambiguous, self-contradictory, or not-verifiable-as-written instruction---is a query-generation fault, so the offending task description is returned to query generation for revision; a \emph{verification-rule defect}---a rule that failed a correctly completed trajectory---is an evaluation fault, so the rule is corrected and the mis-judged trajectory re-enters the evaluation queue. Neither is charged to the model. A well-posed task that merely exceeds the \emph{current} model's capability is not a Level-2 defect---it is handled at Level 3 by RQA.
  \item \textbf{Level 3: model-failure attribution (algorithm side)}. Only after the first two levels are excluded is the failure attributed to the model itself, and further subdivided into structured causes: \emph{perception drift} (the VLM mis-identifies or misses UI elements, including localization offsets and element confusion), \emph{state-tracking errors} (losing context across page transitions, causing looping behavior), and \emph{constraint violations} (failing to follow specific restrictions in the query). The attribution conclusion maps directly onto the ECDM's three processing paths: occasional single-step slips go to step-level correction, well-posed tasks beyond the current model's capability go to Retroactive Query Alignment (RQA), and recurring error classes go to counterfactual query derivation (\S\ref{sec:data-ecdm}).
\end{itemize}

This cascade turns ``evaluation signals'' into ``executable optimization targets'': every level's conclusion has a definite downstream destination---environment features feed environment self-healing, task-specification defects feed query-generation revision and verification-rule defects feed evaluation revision, and model-failure causes feed ECDM repair and the next round of directed data production---so every diagnostic signal has an owner.

\subsubsection{Error Correction \& Derivation Module}
\label{sec:data-ecdm}

The Error Correction \& Derivation Module (ECDM) receives all failed trajectories produced by each round of rollout and evaluation routing. Its task is to turn failure into data. All processing is fully offline pure data engineering, consuming no real-device collection resources. Upon entry, each failed trajectory is dispatched to one of three processing paths by directly consuming the conclusion of the cascaded attribution (\S\ref{sec:data-attribution})---the attribution engine has already determined ``at which level the failure occurred,'' and the ECDM does not re-judge:

\begin{itemize}[leftmargin=1.6em]
  \item \textbf{Occasional single-step slip} (prefix correct; only individual steps went wrong, and the correct action at those steps can be determined offline) $\rightarrow$ step-level correction $\rightarrow$ repair the trajectory $\rightarrow$ SFT pool.
  \item \textbf{Task over-scoped} (trajectory prefix correct, but the original query exceeds capability or contains unreachable subgoals) $\rightarrow$ Retroactive Query Alignment (RQA) $\rightarrow$ downgrade the query to match the prefix $\rightarrow$ SFT pool.
  \item \textbf{Capability gap} (the same error class recurs across multiple trajectories, pointing to a capability deficiency) $\rightarrow$ counterfactual derivation $\rightarrow$ batch-synthesize adversarial queries $\rightarrow$ RL query pool + query generation.
\end{itemize}

The three paths differ in output granularity: the first two repair individual trajectories one by one (producing SFT supervision); the third batch-generates tasks for an entire failure class (producing RL exploration directions).

\paragraph{Step-Level Correction: Repairing Occasional Single-Step Slips.}

When the trajectory prefix is entirely correct and only one step went wrong (tapped the wrong element, accidentally triggered a popup), with the correct action at that step determinable offline without depending on subsequent real-device state, we truncate the first $N-1$ steps as the ``prefix success state'' and repair step $N$ offline through a two-tier mechanism: as the primary track, the erroneous step's context (screenshot, goal, action history) is fed to a stronger offline Teacher model, which rewrites the correct action for that step; as the fallback, cases with low Teacher confidence or hard scenarios (risk control, complex account states, or severe UI deformation) are routed to human repair. \emph{Example.} A trajectory triggers a popup at step~$N$ while steps $1,\dots,N-1$ are correct; we keep that prefix and let the Teacher model rewrite step~$N$ into the correct action, yielding a ``prefix + single-step correction'' label.

Because the real-device state cannot be restored, the repaired action cannot be re-executed and verified on the original trajectory; this path therefore produces a ``prefix trajectory + single-step correction label,'' not a complete post-recovery trajectory. Its supervision value lies in letting the model learn the correct action within the real erroneous-state context of ``having already gone off track''---single-step supervision on ``how to correct after a mistake.'' A complete multi-step ``deviate--diagnose--recover'' trajectory cannot be constructed offline and can only be re-obtained on real devices in the next round of rollout.

\paragraph{Retroactive Query Alignment (RQA): Downgrading an Over-Scoped Task into the Prefix's Correct Solution.}

Sometimes the trajectory prefix is entirely correct, yet the root cause of failure is not ``a wrong step''---the original query was over-scoped, exceeding the current model's capability or containing unreachable subgoals. The prefix is not wrong; what is wrong is the goal it serves. RQA adopts a ``subtract only, never redo'' offline extraction strategy, retroactively downgrading the over-scoped query into a subtask that precisely fits the completed prefix. When the trajectory is detected to deviate from the goal or hit an environment error at step $t$, the actions $a_t$ and beyond are dropped, keeping the entirely correct sequence $a_1, \dots, a_{t-1}$ as the ``golden prefix sub-trajectory''; the screenshot $s_{t-1}$ and layout tree of the last frame before the error are extracted to parse the actual page state; and an offline Teacher model rewrites the original instruction $Q$ into a new instruction $Q'$ that precisely matches the terminal-state semantics of $s_{t-1}$. The reconstructed pair $(Q', [a_1, \dots, a_{t-1}])$ is then injected into the SFT pool as a fully correct short-chain sample, with zero resampling cost.

\emph{Example.} The original instruction $Q$ is ``search for the Sony XM5 headphones on Taobao, also open JD.com to compare prices, and place the order on whichever platform is cheaper.'' The model completed the Taobao search, the JD.com search, and the price comparison, but failed at final checkout because no payment method was bound. RQA truncates the checkout segment, keeps the ``dual-platform search + price comparison'' prefix, and rewrites $Q$ into $Q'$: ``search for the Sony XM5 headphones on Taobao and JD.com respectively, compare prices across the two platforms, and note which one is cheaper.''

In short, step-level correction assumes ``the task is right; one step went wrong,'' so it changes the action; RQA assumes ``the actions are all right; the task was over-scoped,'' so it changes the task.

\paragraph{Counterfactual Query Derivation: Turning a Class of Failures into Targeted Training Tasks.}

The first two paths handle individual trajectories case by case, but a capability gap cannot be fixed by case-by-case repair. After clustering failed trajectories by structured attribution (\S\ref{sec:data-attribution}), if the same error class recurs across multiple trajectories and multiple queries---e.g., ``multi-constraint filtering settled early with only some conditions satisfied,'' ``information extracted on a previous page forgotten after cross-page operations,'' ``tapped Cancel on a system confirmation dialog''---what it points to is a capability deficiency of the model (corresponding to a capability-profile dimension in \S\ref{sec:data-capability-profile}), and systematic reinforcement is needed. The ECDM first induces the structured features of the failure class (triggering scenarios, target operation flows, the missing interaction intuition); the Teacher model then batch-synthesizes adversarial new queries that precisely target the gap based on the failure pattern. \emph{Example.} For ``missing conditions in multi-constraint filtering,'' it generates ``find a restaurant rated above 4.5, with per-person spend under 100 CNY, and delivery support---all three conditions are mandatory''; for ``cross-page forgetting,'' ``remember the price at the first store, compare it at the third store, and order from the cheapest.'' This batch of adversarial queries is injected into the next round's RL query pool and simultaneously fed back into query generation, forcing the model to repeatedly explore at its fragile capability boundary until it converges to the correct policy.

Step-level correction and RQA recover usable supervision from failed trajectories, but only for existing capabilities; counterfactual derivation turns failure patterns to advantage, producing new tasks that attack the gaps---the former repairs data, the latter generates direction.

\subsubsection{Lifecycle of the Self-Evolution Loop}
\label{sec:data-lifecycle}

The self-evolution loop is not endless mechanical repetition. As iterations proceed, the loop exhibits qualitatively distinguishable characteristics across lifecycle stages, with model capability and corpus quality rising steadily together:

\begin{itemize}[leftmargin=1.6em]
  \item \textbf{Early stage $\cdot$ Behavioral cloning and norm establishment}: SFT-dominant. Through the high-purity gold-standard data produced by HTSC, the model rapidly absorbs GUI fundamentals---layout, element localization, and basic interaction norms---building basic task-execution intuition.
  \item \textbf{Middle stage $\cdot$ Exploration-boundary widening and robustness strengthening}: the RL share rises significantly. Large-scale real-device rollouts shape fault tolerance through dynamic rewards; the ECDM batch-revives failed trajectories (step-level correction + RQA) and generates adversarial derived queries at high frequency, feeding back into query generation and forcing the model to self-calibrate in a variable, noisy real-device environment. The model's anti-interference and anomaly self-healing abilities improve markedly in this stage.
  \item \textbf{Late stage $\cdot$ Dynamic balance and self-refinement}: queries corresponding to simple tasks, having sustained high pass@K or reward, are successively removed from the RL active set as mastered samples, and the RL query pool automatically concentrates into the model's hardest capability boundary. Every time the model conquers a high-difficulty long-tail case, the overall difficulty of the corpus shifts up once---forming a self-evolution loop of ``stronger model $\rightarrow$ deeper boundaries discovered $\rightarrow$ harder derived corpus $\rightarrow$ even stronger model.''
\end{itemize}

%% file: sections/model_training.tex
\section{Model Training}
\label{sec:training}

\begin{figure}[t]
    \centering
    \includegraphics[width=0.99\linewidth]{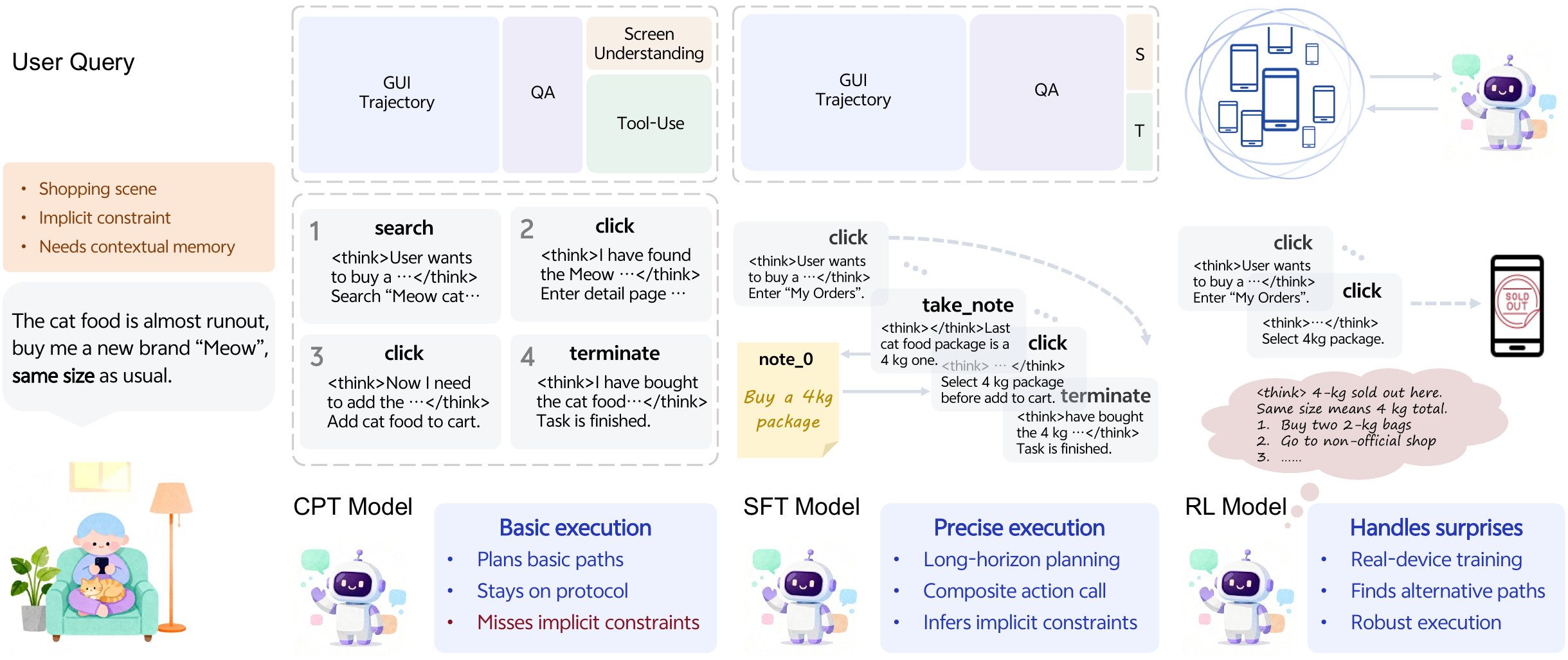}
    \caption{The three training stages of \OurModel. CPT learns multi-step planning over a sequence of screens. SFT learns to follow implicit constraints and call composite actions. RL trains on real devices, gaining robustness and learning to handle real-world surprises.}
    \label{fig:pipeline}
\end{figure}

We train \OurModel from a general VLM in three stages. Continual pre-training (CPT) uses a large-scale uncurated training set and teaches the model to act over a sequence of screens. Supervised fine-tuning (SFT) uses expert-verified trajectories to make each decision accurate. Both stages consume samples from our data generation pipeline. 
Reinforcement learning (RL) then releases the model from imitation: guided by the principle \emph{Every Rollout Is Real}, 
every exploration step runs in an online environment and receives a real response, so the capability the model acquires naturally fits deployment. 
After each rollout, the trajectory flows back into the data pool, turning the flywheel. Fig.~\ref{fig:pipeline} summarizes the three training stages and the capabilities each stage is designed to impart.

\subsection{Continual Pre-training}
\label{sec:cpt}

Base models nowadays understand screen elements well, but they are weak at predicting the next action in a trajectory, where a decision needs the model to track state across steps. What they lack is not knowledge about interfaces but a prior over sequential decisions: which stage of the task is active, whether the last action worked, or whether to move forward or back. 
However, keeping the SFT recipe fixed, SFT on the base model stays well below SFT after CPT (\S\ref{sec:exp_cpt}). CPT and verified data therefore do different jobs. CPT moves a general VLM into the GUI domain, and it does so at a per-sample cost far below that of verified data.

\subsubsection{Training Corpus and Objective}
The CPT corpus holds several million samples. Most of it is GUI trajectories, each a sequence of screenshots paired with the model outputs for those screens; \S\ref{sec:data} describes how we collect and filter them. Three smaller sets protect abilities the base model already has. Screen understanding data keeps grounding accuracy from drifting. Tool-use data helps because GUI task operates in a similar way. 
We also introduce general question-answering and reasoning data, which keeps the reasoning ability that goal decomposition and error handling depend on.

We use standard autoregressive language modeling, update all language-model parameters and freeze the vision encoder and the projector.
Let $y_t$ be the response at step $t$ and $x_t$ be its input context, 
the loss is calculated in the following function:
\begin{equation}
    \label{eq:cpt_loss}
    \mathcal{L}_{\mathrm{CPT}}(\theta)
    = -\,\mathbb{E}_{(x_t,y_t)\sim\mathcal{D}_{\mathrm{CPT}}}\!\left[\sum_{j=1}^{|y_t|}
      \log \pi_\theta\!\left(y_t^{j}\,\middle|\, y_t^{<j},\, x_t\right)\right].
\end{equation}

\subsubsection{Interaction Protocol}
This stage also fixes how the model talks to the environment: one input format, one output format, one action space. Later stages reuse all these three. The input at step $t$ has four parts: a system \textbf{p}rompt $\mathcal{P}$ that lists the action space and the rules to follow, the user \textbf{g}oal $g$ in natural language, the interaction \textbf{h}istory $\mathcal{H}_t=\{(c_i,a_i)\}_{i=1}^{t-1}$, and the visual \textbf{o}bservations $o_t$ of the screen. The output at each step is a text segment $c_t$, holding the reasoning trace and a summary of progress, plus one structured \textbf{a}ction $a_t\in\mathcal{A}$ that the device executes:
\begin{equation}
    \label{eq:io}
    (c_t,\,a_t)\;\sim\;\pi_\theta\!\left(\,\cdot\;\middle|\;
    \mathcal{P},\; g,\; \mathcal{H}_t,\; o_{\max(1,\,t-K+1):t}\right),
\end{equation}
where $o_t$ is the screen after $a_{t-1}$ ran and before we decide $a_t$, and $K$ is the size of the visual context window. We treat the two kinds of memory differently on purpose. We keep all past text, which holds the model to the same goal over a long task. We keep only the last $K$ screenshots, which keeps context length and inference cost affordable. 

The action space $\mathcal{A}$ has fourteen actions in four groups (Tab.~\ref{tab:action_space}). 
The first two groups are primitive actions, split by whether a touch is discrete or sustained: tap-level actions are single-point touches and key presses, while gesture-level actions involve continuous contact like swiping and dragging.
The third group contains composite capabilities, each of which folds a long primitive sequence into a single intent-level call. The fourth group covers control and handoff: actions that pause the trajectory, end it, or hand authority back to the user. Putting all four groups in one space means the model does not choose between two modes; it weighs every option in one decision. After CPT the model follows the protocol reliably and emits legal calls with a stable format across steps.

\begin{table}[t]
    \centering
    \small
    \caption{The action space $\mathcal{A}$: fourteen actions in four groups.}
    \label{tab:action_space}
    \begin{tabular}{@{}p{0.20\linewidth}p{0.20\linewidth}p{0.52\linewidth}@{}}
        \toprule
        \textbf{Group} & \textbf{Action} & \textbf{Behavior} \\
        \midrule
        Tap-level action
            & \texttt{click}            & Tap at coordinate $(x,y)$. \\
            & \texttt{double\_click}    & Tap twice quickly, for zooming or text selection. \\
            & \texttt{type}             & Enter text into the widget at a given coordinate. \\
            & \texttt{system\_button}   & Press a device key such as menu, enter, or screenshot. \\
        \midrule
        Gesture-level action
            & \texttt{swipe}            & Slide between two coordinates to scroll or turn a page. \\
            & \texttt{drag}             & Move an element to a target coordinate and release. \\
            & \texttt{long\_press}      & Hold to open a context menu or enter an editing state. \\
        \midrule
        Composite capability
            & \texttt{open}             & Declare a functional need and let the device resolve it to an installed app. \\
            & \texttt{set\_scroll\_widget} & Set a wheel picker to a target value in one call. \\
            & \texttt{take\_notes}      & Save useful screen content to memory; credentials and card numbers are not allowed. \\
        \midrule
        Control and handoff
            & \texttt{wait}             & Skip one step while a page loads. \\
            & \texttt{terminate}        & End the trajectory with success or failure. \\
            & \texttt{clarify}          & Ask a question or confirmation, with optional choices. \\
            & \texttt{call\_user}       & Pause and hand over control, with a reason. \\
        \bottomrule
    \end{tabular}
\end{table}

\subsection{Supervised Fine-tuning}
\label{sec:sft}

After CPT the model makes progress on tasks and respects the protocol. But a legal trajectory is not always the right one or the short one, and the model still breaks rules from time to time. SFT fixes both: at every step we align the model with a verified decision and with the rules it was given. 
We change the data mix to match. GUI trajectories take a larger share, and we draw them only from the high-confidence pool built by cross-model voting and expert verification (\S\ref{sec:data-evaluation}). We cut screen understanding and tool-use data to the smallest amount that keeps those abilities alive. We keep a large share of general reasoning data, because goal decomposition and reading implicit constraints rely on it.
The training setup and objective are the same as in CPT.

\subsubsection{Intent-Level Actions}
The second goal of this stage is to let the model act at the level of intent rather than of primitives. Primitives are easy to learn, but they come with longer sequences, more time on device, and more risk. So we fold common primitive sequences into actions defined by intent. Take launching an app as an example. Many apps can meet the same need, but which ones are installed depends on the phone, the region, and user preference. With primitives only, the model has to try candidates one by one. 
We therefore let \texttt{open} take a description of the need instead of a fixed app name, and the device resolves that description to an installed app. In this way, the model decides what to do next; the device works out how to do it.
And other composite capability actions follow the same pattern.

\subsubsection{Behavioral Boundaries}
If the instruction itself is unclear, 
or if the model reaches a boundary it must not cross on its own, 
the model has to involve the user. Two actions cover these cases. \texttt{clarify} collects what is missing into one question and resumes with the session intact once the user answers. \texttt{call\_user} stops at the critical screen, hands control back, and supplies the information the user needs to decide.

Which action applies depends on what is missing. \texttt{clarify} handles a gap in the instruction, where the model cannot tell which of several readings the user meant. \texttt{call\_user} handles a consequence the model is not entitled to accept on the user's behalf, such as a payment or a deletion. Both actions already exist after CPT, what SFT teaches the model is the timing: when to ask the user, and when to go ahead on its own.


\begin{figure}[t]
    \centering
    \includegraphics[width=0.99\linewidth]{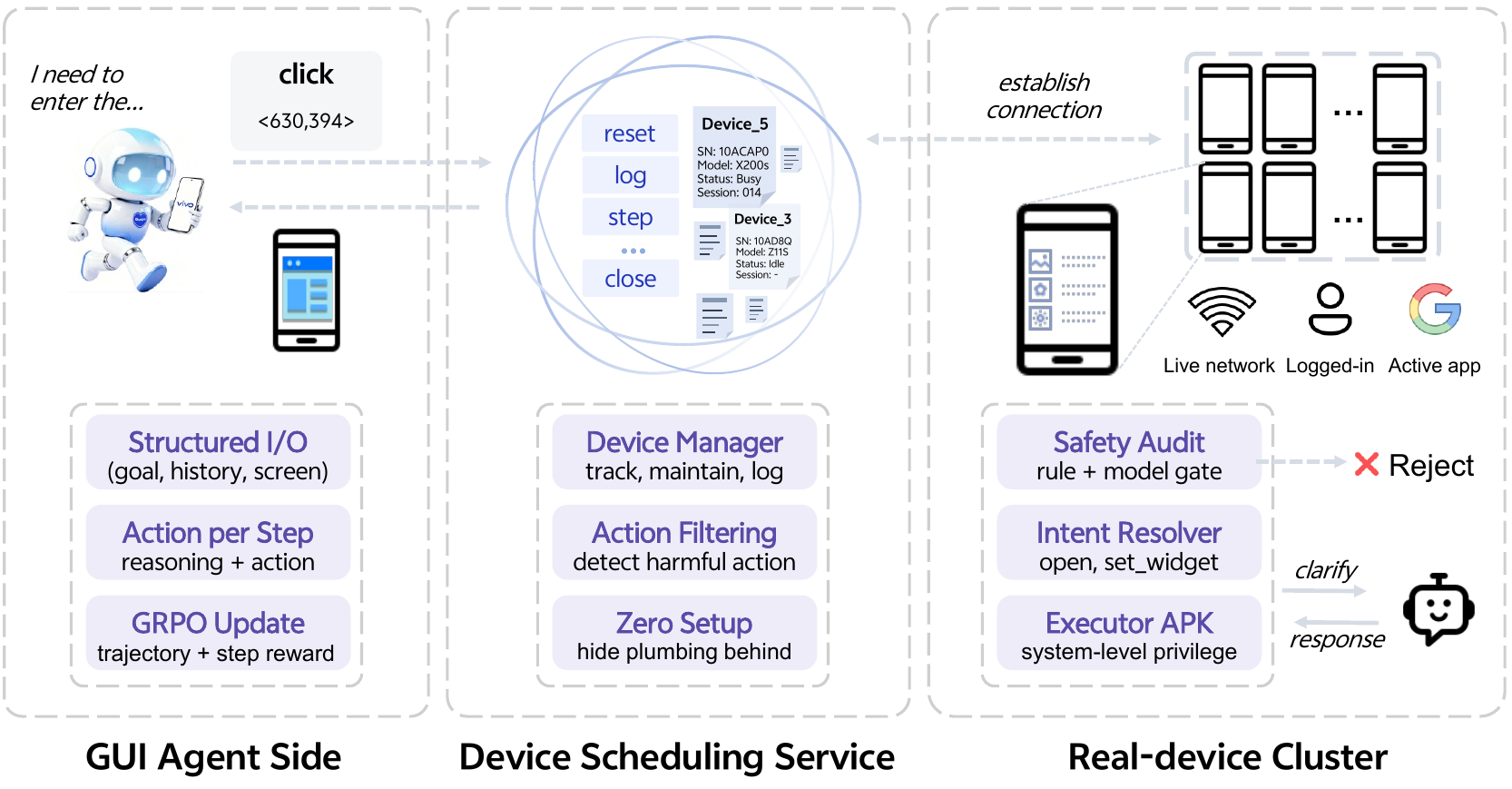}
    \caption{The real-device training environment. A scheduling service gives the trainer a standard RL interface over a cluster of real devices.}
    \label{fig:env}
\end{figure}

\subsection{Agentic Reinforcement Learning}
\label{sec:rl}

SFT can only teach the model to reproduce the trajectories we annotated, and the annotations can cover only a fraction of the obstacles a real device will throw at it. Learning to recover from unforeseen obstacles requires the model to try its own solutions and see how they turn out.
This rules out scoring a fixed set of collected trajectories, because a GUI task usually has multiple valid paths and a reward defined on a single one of them would reject the others. We therefore perform online training: 
the model acts in an online environment, and the reward is computed from 
the actual resulting state.

\subsubsection{Real-Device Environment}
Our environment has three layers (Fig.~\ref{fig:env}). The model emits actions in the form of Eq.~\ref{eq:io}. 
A scheduling service exposes the usual reinforcement learning interface of reset, step, and close, and hides everything below it, so the trainer can rollout through many phones at once. 
The bottom layer is a cluster of hundreds of phones of different models, each running a resident runtime that executes an action and returns the new screen.

The phones in our cluster carry up-to-date apps, live network, and signed-in accounts, so the model meets real pop-ups and real loading delays during training rather than after release.
However, on the other hand, 
free exploration on such devices is risky, so every action passes through a safety audit system that blocks actions involving private data, money, or prohibited content.
This system is the one we ship in production, so the same safeguard that protects our devices during training stays in place after release. 
The phones also run a local harness that expands composite capability actions into concrete operations, keeping the action space identical across training and deployment.

\subsubsection{Reward Design}
We train on tasks from the middle difficulty band of our data engine, since tasks that every model solves or every model fails produce no useful gradient. 
An LLM judge scores each trajectory $\tau$ from the goal, the full sequence of outputs and actions, and the screen at every step, so the score reflects both the decisions and their effect on the device. We query the judge multiple times and take a majority vote. The trajectory reward then combines \emph{progress} and \emph{quality} at equal weight:
\begin{equation}
    \label{eq:traj_reward}
    R(\tau)\;=\;\alpha\cdot\underbrace{\frac{N_{\mathrm{done}}}{N_{\mathrm{nec}}}}_{\text{progress}}
    \;+\;(1-\alpha)\cdot\underbrace{\frac{N_{\mathrm{nec}}}{T}}_{\text{quality}},
    \qquad \alpha=0.5,
\end{equation}
Here $N_{\mathrm{nec}}$ is the number of sub-goals the judge considers necessary for the task, $N_{\mathrm{done}}$ is how many of them the trajectory reached, and $T$ is the number of steps it took. Progress rewards the model for getting as far as it can, while quality measures how many of its steps were actually needed, which prevents it from reaching the goal through a long detour.

A single reward per trajectory tells the model that an attempt failed but cannot tell which step $t$ was responsible for it. We therefore also ask the judge to rate every action,
\begin{equation}
    \label{eq:step_reward}
    r_t \;=\; \mathrm{Judge}\big(g,\;\tau,\;t\big)\;\in\;\{+1,\,0,\,-1\},
\end{equation}
where the three values correspond to a step that is necessary, a step that is harmless but useless, and a step that pushes the task off course.

\subsubsection{Training Optimization}
We optimize with Group Relative Policy Optimization (GRPO). For each task we sample $G$ trajectories, normalize the two rewards separately, and add them into a single advantage:
\begin{equation}
    \label{eq:advantage}
    A_{i,t}\;=\;\frac{R(\tau_i)-\mathrm{mean}\big(\{R(\tau_j)\}_{j=1}^{G}\big)}
                     {\mathrm{std}\big(\{R(\tau_j)\}_{j=1}^{G}\big)}
    \;+\;\lambda\cdot\frac{r_{i,t}-\mathrm{mean}\big(\{r_{i,t'}\}_{t'=1}^{T_i}\big)}
                          {\mathrm{std}\big(\{r_{i,t'}\}_{t'=1}^{T_i}\big)}.
\end{equation}
The trajectory reward is normalized within the group of one task, where it compares whole solutions against each other, and the step reward is normalized within one trajectory, where it compares individual steps. The weight $\lambda$ balances the two. The policy objective is
\begin{equation}
    \label{eq:grpo}
    \mathcal{J}(\theta)=\mathbb{E}_{\tau_i\sim\pi_{\theta_{\mathrm{old}}}}\!\left[\frac{1}{G}\sum_{i=1}^{G}\frac{1}{T_i}\sum_{t=1}^{T_i}
    \min\Big(\rho_{i,t}A_{i,t},\;
    \mathrm{clip}\big(\rho_{i,t},1-\epsilon,1+\epsilon\big)A_{i,t}\Big)\right],
\end{equation}
where $\rho_{i,t}$ is the probability ratio between the current and the sampling policy. We also add a KL term toward the SFT model, which limits how far exploration can move the policy and protects the protocol and the safety boundaries learned earlier.

Additionally, we adjust the rollout schedule during training: 
groups whose success rate falls outside a useful range are discarded because they carry no contrast, and when sampling fails we fall back on earlier successful trajectories of the same task, with a cap on how much of an update they may contribute. 
Every trajectory we collect is kept and returned to the data pool, where verified successes become supervision for later runs and failures are routed through cascaded attribution into the ECDM. This closes the loop with the data engine of \S\ref{sec:data-self-evolution}, and lets the model keep improving across iterations instead of finishing in one pass.

%% file: sections/vivo_benchmark.tex
\section{\BenchPro}
\label{sec:benchmark}

\emph{Every Query Evolves.} A fixed benchmark saturates as \OurModel improves and the scores stop telling us what to fix. What we need is therefore a benchmark that can evolve under a fixed metric system. Our three axes---scenario, complexity, and interaction \& risk---serve as a stable coordinate system. When one benchmark saturates, we shift the quotas along a few dimensions and obtain a new one that differs from the old only in an explicitly stated way, so every score change is attributable. 
Our construction pipeline consists of five steps.
First, we define a set of annotation dimensions that turns the question of what to test into a question of how to allocate quotas (\S\ref{sec:metric}). 
Second, we write the purpose of a new benchmark as a target quota over those dimensions. 
Third, an LLM generates queries against the quota.
Fourth, experts verify and refine every query before it ships (\S\ref{sec:bench_construction}). 
Fifth, we score each trajectory through independent multi-judge voting (\S\ref{sec:bench_eval}).

\begin{figure}[h]
    \centering
    \includegraphics[width=0.99\linewidth]{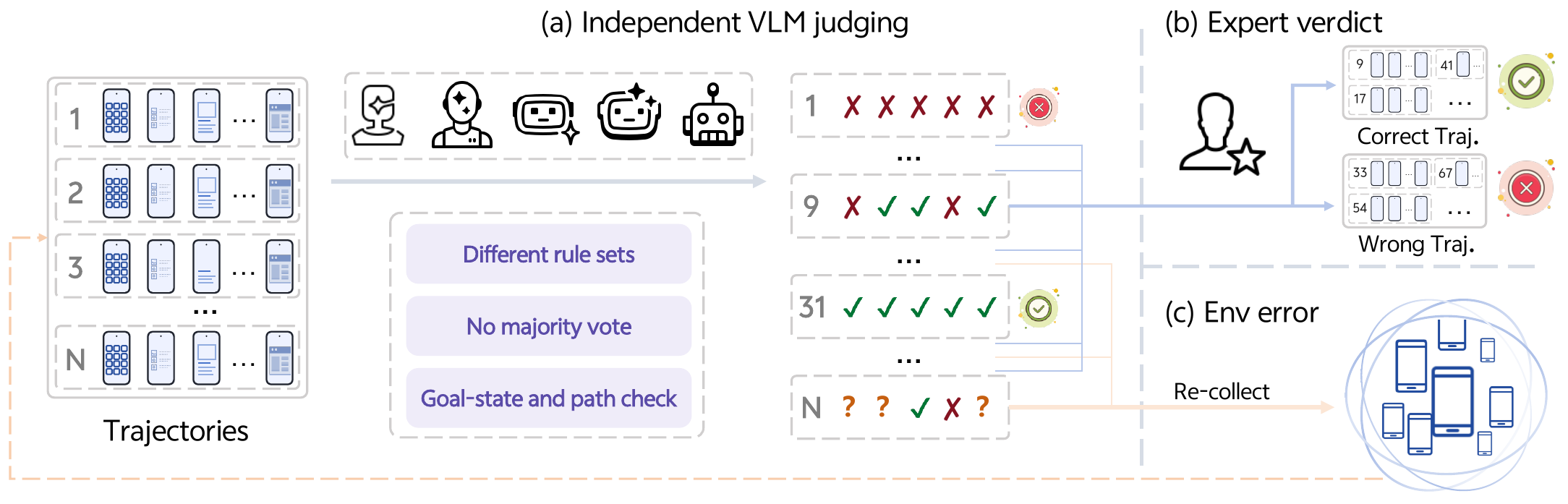}
    \caption{The evaluation protocol. A trajectory is scored by multiple independent LLM judges. When all judges agree, the verdict stands; when they split, the case goes to an expert verdict. Trajectories broken by the environment are re-collected.}
    \label{fig:benchmark-evaluation}
\end{figure}

\subsection{A Quantified Metric System}
\label{sec:metric}

Our annotation dimensions exist so that queries can be produced in bulk to a specification. Tab.~\ref{tab:metric_system} lists them. They form three orthogonal axes, which ask where the task happens, how hard the execution is, and what form the interaction takes. 
Every value is a discrete level, so any benchmark could be defined by one set of quotas over this space. 

\subsubsection{Scenario Metric}
The top-level split is between system apps and third-party apps, and it reflects two demands we cannot merge. Our goal is the end-user experience, where a task in Settings or Gallery matters as much as one in a shopping or navigation app. 
However, comparing models on system apps would nevertheless be unfair, 
since system apps are designed differently.
Everything we report and compare in this work therefore uses third-party apps. 

Below this split we spread queries over twelve app categories, such as shopping, video, and messaging, so that no small group of apps dominates the benchmark. 
The finest level is the functional entry point. Two functions inside one app might differ more than two apps do, and a quota defined only over apps tends to keep hitting shallow functions on the main pages of apps. We therefore run an automated discovery of in-app functionalities and use the entry points it finds as our candidate pool. 
Coverage then follows the functions an app actually offers, not how well the query author happens to know that app.

\subsubsection{Complexity Metric}
The main measure here is path complexity, which we define by the number of GUI steps a task needs and record as simple, moderate, or complex. One constraint decides whether this label is worth anything: the step count cannot be derived from the wording of the instruction. \emph{Book me a ticket} is four words long and can take more than ten steps on a real interface. 
We therefore have an LLM walk through the likely path for each query, and we correct its estimates during expert refinement.

Orthogonal to path complexity is intent dependency, with three values: single-intent, independent multi-intent ($A + B$), and dependent multi-intent ($A \rightarrow B$). 
In independent multi-intent ($A + B$), a query bundles several sub-tasks that can be completed in any order without passing results between them. 
And in dependent multi-intent ($A \rightarrow B$), the model must feed the output of one sub-task into the next, and a query can demand this without being long.
We also record app scope, which is single-app or cross-app, and operation type, which is tap-level or gesture-level.

\subsubsection{Interaction and Risk Metric}
The first two axes assume that a task should be carried out. But real use does not work all that way, and this axis covers what lies outside plain execution. We set instruction explicitness that has three values: an explicit instruction, an implicit intent that names a function but leaves out the path, and an underspecified instruction that states a problem without a goal. 
Instruction type separates agent-executable queries from agent-user interactive ones, where the model is supposed to stop at a decision point and ask rather than decide for the user. 
Refusal and safety covers requests an app cannot support: requests beyond the capability scope of the agent, commonsense-violating requests, and high-sensitivity operations. For these queries the correct behavior is to refuse or to clarify. Without them, a model that acts regardless of the request would score higher than one that knows its limits.

\begin{table}[t]
    \centering
    \small
    \caption{Our metric system. Three orthogonal axes cover where a task happens, how hard it is to execute, and what form the interaction takes. All values are discrete, so a benchmark is defined by a set of quotas over this space.}
    \label{tab:metric_system}
    \begin{tabular}{@{}p{0.16\linewidth}p{0.19\linewidth}p{0.28\linewidth}p{0.29\linewidth}@{}}
        \toprule
        \textbf{Axis} & \textbf{Dimension} & \textbf{Values} & \textbf{What it tests} \\
        \midrule
        Scenario
            & App ownership        & system / third-party & On-device experience; fair comparison across models \\
            & App category         & 12 categories, e.g.\ shopping, video, messaging & Generalization across scenarios \\
            & Functional intent & varying across apps & Reachability of deep functions \\
        \midrule
        Complexity
            & Path complexity      & simple $\leq 5$ / moderate $6\text{--}10$ / complex $>10$ steps & Long-horizon execution and progress tracking \\
            & Intent dependency    & single / independent ($A+B$) / dependent ($A \rightarrow B$) & Task decomposition and passing results across steps \\
            & App scope & single-app / cross-app & Cross-app state transfer \\
            & Operation type & tap-level / gesture-level & Fine-grained gestures \\
        \midrule
        Interaction \& risk
            & Explicitness & explicit / implicit / underspecified & Intent understanding and commonsense completion \\
            & Instruction type     & agent-executable / agent--user interactive & When to ask instead of deciding for the user \\
            & Refusal and safety   & unsupported / beyond scope / non-GUI / high-sensitivity & Capability-boundary awareness and risk avoidance \\
        \bottomrule
    \end{tabular}
\end{table}

\subsection{Benchmark Construction}
\label{sec:bench_construction}

\begin{figure}[t]
    \centering
    \includegraphics[width=0.95\linewidth]{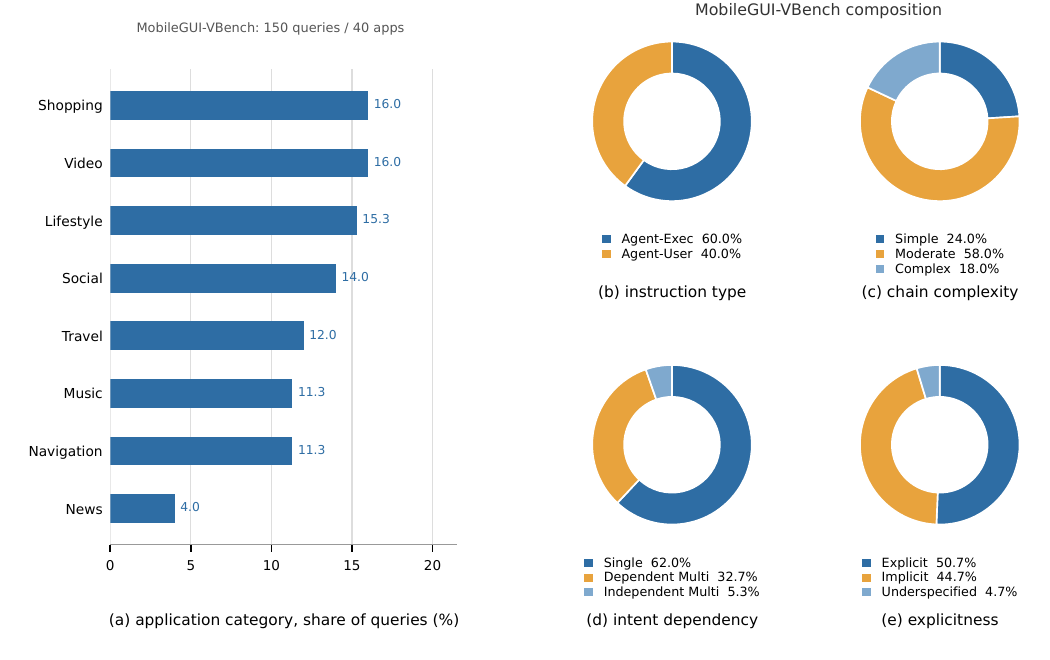}
    \caption{Composition of \BenchPro. The bar chart gives the distribution over app categories, and the rings give the distribution over path complexity, intent dependency, instruction explicitness, and instruction type.}
    \label{fig:benchmark}
\end{figure}

The first output of construction is a distribution table. 
It states which direction generation should lean and which scenarios may not be missing, and it lets two benchmarks built on the same dimensions be compared directly. We built two. \BenchBasic answers whether the model is usable at all,
and \BenchPro answers where the model still falls short. It holds 150 queries over 40 apps, and its quota is not a scaled-up copy of \BenchBasic but a deliberate skew along a few directions. Fig.~\ref{fig:benchmark} shows the resulting composition.

\subsubsection{Scenario Quotas}
\BenchBasic has to answer whether the model works on the phone a user actually holds, so its scenario quota follows app popularity rather than an even grid. 
Its current version spans 43 apps, 34 third-party and 9 system ones, which together cover 36 of the 50 most installed titles. The system apps are universal utilities such as Messaging, Gallery, and Calculator.
Eleven categories are represented, five of which take 12 to 16 percent each while the rest form a long tail.

With the sole exception of the system Gallery, \BenchPro uses only third-party apps, keeping scores fair to compare across models. 
Within that scope we keep the major categories, such as shopping and video, at similar quota, and we retain a few long-tail categories such as news reading, so that no single interface style drives the conclusions. 
We also cap how many queries any one app may contribute, so that the implementation quality of one app cannot hold the whole score hostage.

\subsubsection{Complexity Quotas}
\BenchBasic allots path complexity roughly as $3:4:2$ over simple, moderate, and complex. 
What the quota spends least on is the combination of a short path and a rarely used function, since a benchmark meant to certify usability learns little from it. 

\BenchPro keeps moderate paths as the main body but cuts simple ones further, retaining only a few as a lower anchor, since they no longer separate models. Its added difficulty comes less from longer paths than from intent dependency, whose dependent multi-intent share we raised sharply. 
We weighted this direction because it tests directly whether the model can pass the result of one sub-task into the next. We also raised the share of gesture-level operations somewhat, while keeping tap-level operations in the majority, as they remain the common case in real use.

\subsubsection{Interaction and Risk Quotas}
\BenchBasic sets no quota for refusal and safety. Its agent--user interactive queries already test when the model should pause and ask rather than decide, but the benchmark never asks it to refuse an operation. A model that never says no could therefore score well on \BenchBasic while being reckless in practice. \BenchPro was built to close this gap.

The other structural change in \BenchPro is that agent-user interactive queries now make up close to 40 percent of the benchmark. On a real phone, deciding for the user at a point that needed confirmation is usually worse than pausing to ask. Giving these queries a large share rewrites the standard of success from whether the operation was completed to whether the model responded appropriately at the right moment. 
Risk scenarios follow a different rule: rather than getting a proportional share, they must simply be present. 
Queries that touch money, private data, or any irreversible change belong in the benchmark however small their share, because otherwise a model that acts regardless of the request scores higher than it should. 
We did not invent these quotas either. We collected failure cases from our own models, picked out the scenarios they got wrong, and generalized from those, so that the skew points at real weaknesses.

\subsubsection{From Quotas to Queries}
Once the quota table is fixed, we turn the dimension definitions and the annotation rules into structured constraints, and an LLM generates queries against the target distribution while annotating them at the same time. 
We validate every field as it is produced, then compare the realized distribution against the quota and patch the parts that drift. What the model produces is a draft. 
Before a benchmark ships, experts go through it query by query: they rewrite stiff phrasing into what a user would actually say, correct wrong labels, and attach to each query a reproducible description of the environment it assumes, marking whether that condition holds once or has to be reset for every run. This last item is what makes results comparable at all. Unless we know which parts of the device state must be restored, the same query run twice, or run on two models, is not the same query.

\subsection{Evaluation Protocol}
\label{sec:bench_eval}

One evaluation of a GUI agent is a trajectory that actually happens on a phone, so we report a single metric: the success rate, the fraction of queries on which the model completes the whole task by itself. Writing $\mathcal{Q}$ for the benchmark and $\tau_q$ for the trajectory the model produces on query $q$,
\begin{equation}
    \label{eq:sr}
    \mathrm{SR} \;=\; \frac{1}{|\mathcal{Q}|}
    \sum_{q \in \mathcal{Q}} \mathbb{1}\!\left[\,\mathrm{Judge}(q,\,\tau_q) = \mathrm{success}\,\right],
\end{equation}
where $\mathrm{Judge}$ returns a binary verdict and we award no partial credit. We give no credit for partial paths because on a real device getting most of the way there is worth nothing to the user. The same rule decides the other two kinds of query. On an agent--user interactive query, a model that passes a decision point involving money or private data without calling \texttt{call\_user} fails, even if the task appears to have been completed. On a refusal query, refusing or clarifying correctly counts as success. A single metric thus constrains execution and interaction behavior at once, and a model cannot trade recklessness for score.

Judging is the hard part, because deciding whether a trajectory reached its goal usually means reading the semantics of the last screen and often of several screens before it. 
We use several VLM judges, fine-tuned for this task, with different rule sets, each ruling independently on whether the goal state was reached and whether the model interacted at the right moments, and we count a trajectory as successful only when all of them agree.
This unanimity requirement is deliberate, because it turns disagreement among judges into a signal: a trajectory the judges split on lies outside the range where automatic judging is reliable. We export those cases for experts to score by hand, and we use their verdicts to fix whatever was ambiguous in the judging rules.

Real-device evaluation also suffers from failures that have nothing to do with the model, such as an expired login, a redesigned interface, or a network error. Counting these as failures would mix unrelated variance into the metric. We therefore attribute every failure first, and a trajectory broken by the environment is not scored. We reset the device and collect it again.

%% file: sections/experiments.tex
\section{Experiments}
\label{sec:exp}

\subsection{Training Details}
\label{sec:exp_details}

All models start from the same base model, Qwen3.5-35B-A3B, and pass through CPT, SFT, and RL in that order, with each stage continuing from the output of the previous one.

\paragraph{Continual pre-training.}
We train our model in the CPT stage on 64 GPUs.
The training corpus, consisting of 15B tokens, draws on four complementary sources, roughly 70 percent is GUI trajectories data, others include screen understanding, question answering, and public tool-use data.
Both GUI and screen data are collected from real devices.
We freeze the vision encoder and the projector and update only the language model, so that the model picks up a temporal decision prior while keeping the perception ability of the base model. The learning rate is $1\times10^{-5}$ with a warmup fraction of $0.05$ and cosine decay to $1\times10^{-6}$, the batch size is 128, the maximum sequence length is 10240, and we train for one epoch.

\paragraph{Supervised fine-tuning.}
SFT training uses the same freezing strategy on 32 GPUs and is conducted on a 5B-token corpus.
GUI trajectories at this stage are collected on real devices and verified by experts, so the corpus is smaller than the CPT one but far higher in annotation quality and task complexity. 
The learning rate and the schedule follow CPT, the batch size rises to 128, the maximum sequence length is 10240, and we train for one epoch.

\paragraph{Reinforcement learning.}
RL is optimized via GRPO on 32 GPUs, interacting with an environment consisting of hundreds of real devices and Android emulators.
The training instruction set holds over a thousand queries. 
Each iteration samples 8 instructions and rolls out 8 trajectories per instruction, and these 8 trajectories form the group over which we compute advantages. 
The weight $\lambda$ is set to 1 by default.
The learning rate drops to $1\times10^{-6}$ and the batch contains 512 interaction windows. A trajectory runs for at most 20 interaction steps, visual context window keeps most recent 5 frames, and the prompt is capped at 12288 tokens.

\subsection{Comparison with State-of-the-Art Models}
\label{sec:exp_sota}

\input{tables/main_table}

We compare \OurModel against the strongest agents we could reach, on two benchmarks: \BenchPro, and the public AndroidWorld. The baselines fall into three groups. Closed-source models reached through an API, closed-source models reported by their authors but not released, and open-source models. Every number in Tab.~\ref{tab:gui_only} is a success rate on a real device or in a live environment, so the comparison covers whole trajectories rather than single-step predictions.

\OurModel reaches 87.4 on \BenchPro with 35B total and 3B activated parameters. The best closed-source API model scores 82.3, and the best open-source baseline scores 66.4. We read this gap as a benchmark that tests what our metric system was built to test: most of the \BenchPro queries carry dependent multi-intent structure, or ask the model to stop and consult the user, or must be refused outright. 
A model tuned for straightforward execution has nowhere to earn those points. The spread within the baselines supports this. Several models that do well on AndroidWorld drop sharply here, and PhoneBuddy is the clearest case, at 83.2 on AndroidWorld against 39.0 on \BenchPro.

On AndroidWorld, \OurModel reaches 84.9, surpassing all open-source baselines and closed-source models, behind only Qwen3.8-Max (85.3) and Qwen3.8-Flash (84.5) — both of which activate far more parameters than we do. What we want to point out is not the ranking but the shape of it. The models ahead of us on AndroidWorld are 64.2 and 59.1 on \BenchPro, roughly 25 points below us, while we sit within 2 points of them on AndroidWorld. Strong execution therefore does not carry over to the interaction and refusal behavior that \BenchPro asks for, whereas a model trained for the latter keeps its execution ability intact.

\subsection{Ablation Studies}
\label{sec:exp_ablation}

\subsubsection{The Necessity of Continual Pre-training}
\label{sec:exp_cpt}

\input{tables/tab_cpt}

To test whether CPT is needed at all, we fix the SFT configuration and change only what comes before it: whether the model first sees the large weakly supervised corpus. We evaluate on screen understanding and on trajectory execution. The first measures single-step grounding and semantics of interface elements, the second measures multi-step execution by success rate.

Tab.~\ref{tab:cpt} shows that the base model already scores 95.48 and 94.34 on screen understanding, yet completes only 36.6 of the trajectories, even though its prompt contains examples of how to call actions. 
The bottleneck is therefore not reading the screen. What the base model lacks is a notion of the action space, a causal link between consecutive steps, and any tracking of how far the task has progressed. 
SFT raises the success rate to 64.1 while leaving screen understanding flat, which tells us that expert-verified trajectories supply exactly this missing decision ability. Adding CPT on top pushes the success rate to 70.8, at the cost of slight degradation on the screen understanding benchmarks. 
Since both variants use identical SFT data, those 6.7 points come from the prior that CPT installs. When annotation budget is the binding constraint, building that prior from a large weakly supervised corpus is a path to improvement that does not require more annotation.

\subsubsection{The Form of the Interaction Context}
\label{sec:exp_context}

\begin{figure}[t]
    \centering
    \includegraphics[width=0.8\linewidth]{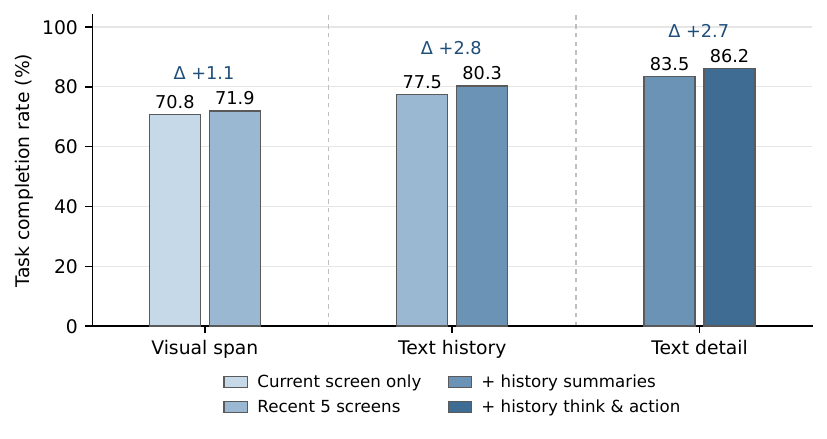}
    \caption{Extending the interaction context raises the success rate. The three pairs sit on different baselines, so only within-pair differences are comparable.}
    \label{fig:context}
\end{figure}

We next ask how the history should be organized for the model to use it. We extend what the model can see along two directions, visual and textual, in four settings: the current screen only, the five most recent screens, five screens plus the summaries of earlier steps, and five screens plus the full reasoning traces, actions, and summaries of earlier steps. 
Training and evaluation cost kept us from placing all four on one baseline. We instead ran three pairwise comparisons, each of which changes only the context form and holds everything else fixed, so the conclusions hold within a pair and should not be read across pairs.

The three pairs in Fig.~\ref{fig:context} give $+1.1$, $+2.8$, and $+2.7$ points of success rate, so extending the history helps in every direction, but the visual and textual extensions are not worth the same. 
Going from one frame to five frames gives the smallest gain, $+1.1$. Earlier screens mostly supply continuity of appearance, while the information a decision actually needs, namely how far the task has gone and what the model has already tried, is hard to recover reliably from pixels. 
Stacking raw observations therefore does not give the agent the information it needs.
The two textual extensions give about $+2.8$ each, both clearly above the visual side, so the model relies on history that has already been written down in language. The third pair also shows that reasoning traces and actions carry information the summaries alone do not, which is why we keep all three in the context rather than summaries only.

\subsubsection{The Data Mixture across Stages}
\label{sec:exp_ratio}

\input{tables/tab_ratio}

The base model is already strong enough at screen understanding tasks but weak at multi-step execution, which suggests that the mixture should tilt toward the latter. We test this by changing only the mixture, holding the data sources and the training procedure fixed: we cut the screen understanding data, which brings the total CPT corpus down by about 40\%, and at the same time raise the trajectory share by a factor of four in CPT and two in SFT.

Tab.~\ref{tab:ratio} shows the success rate rising from 70.8 to 71.9, while the two screen understanding scores move in opposite directions and stay flat overall. The gain arrives with a corpus 40\% smaller, which is a win on both counts: we train faster and still execute better. 
Screen understanding data therefore reaches a point beyond which more of it no longer improves execution. The screen understanding scores also survive the cut, which means the base model already learned this ability during its general pre-training, and our training only has to adapt it to the interfaces our target apps use. 
So we design the mixture by verifying which abilities the base model already brings, giving those just enough data to transfer, and spending everything else on the abilities it lacks.

\subsubsection{The Role of the Real-Device Online Environment}
\label{sec:exp_online}

\input{tables/tab_online}

What RL returns depends on where the interaction data comes from. To separate the contribution of the algorithm from that of the environment, we start from one SFT model and apply two RL settings. 
The first optimizes on trajectories collected in advance, where the model's outputs are never executed, so it never sees what its own choices lead to. 
The second samples in the real-device online environment, where every output is actually executed and returns the real next screen. 
Both use the same optimization algorithm, and the only difference is whether the interaction data is replayed or produced online.

Tab.~\ref{tab:online} shows that the offline setting gains nothing over the SFT starting point, while the online setting raises the success rate to 82.0. Applying an RL objective to existing trajectories is therefore not enough. 
The state transitions in an offline trajectory were fixed, so once the current policy drifts away from that distribution, the advantage estimate no longer corresponds to any real consequence, and RL degenerates into reweighting the actions the dataset already contains.
The real-device online environment closes the loop between an action and its consequence. The model sees how its own mistakes change the following state, and it finds workable paths that never appear in the annotated trajectories, which is where error recovery and multi-path solving come from. 
For long-horizon tasks whose state transitions depend heavily on how the environment responds, online interaction is not an optional refinement of RL but a precondition for it.

\subsubsection{The Reliability of the Reward Signal}
\label{sec:exp_reward}

\input{tables/tab_reward}

In online RL the update direction comes entirely from the reward, and a model-based judge is itself stochastic: the same trajectory can receive different verdicts on different passes. 
Group-relative advantages amplify it, because once a trajectory is scored too highly by mistake, the behavior it contains keeps getting reinforced. We therefore rebuilt the judging procedure. 
Instead of one holistic judgment of whether the task was completed, we write a set of atomic verification conditions that can be checked one by one, bind each condition to a designated frame, and require all of them to hold. 
This replaces the earlier disjunctive bypass rules, under which hitting any one condition was enough to pass. We also judge each trajectory five times to suppress single-pass randomness. 

Tab.~\ref{tab:reward} shows the success rate rising from 72.52 to 77.42. Because the two runs share the same reward definition, those 4.9 points do not come from a better objective. They come from measuring the same objective more accurately. 
Online RL is thus limited by the judging step at least as much as by the optimization step. 
When the reward is noisy, sampling more or training longer can amplify a misleading signal, and making the judge more consistent does more good than tuning the algorithm.

%% file: tables/main_table.tex
\begin{table}[t]
\centering
\begin{tabular}{llccc}
\toprule
\textbf{Model} & \textbf{Access} & \textbf{Params} & \textbf{MobileGUI-VBench} & \textbf{AndroidWorld} \\
\midrule
\rowcolor{sectionblue}
\multicolumn{5}{l}{\textit{Closed Source}} \\
Seed2.1-Pro          & API & \faIcon[regular]{lock} & 82.3 & 72.4 \\
Seed2.0-Pro          & API & \faIcon[regular]{lock} & 80.8 & 71.5 \\
Seed2.0-Lite         & API & \faIcon[regular]{lock} & 77.5 & 71.6 \\
Gemini3.7-Flash      & API & \faIcon[regular]{lock} & 78.3 & 80.1 \\
Claude Opus 4.8      & API & \faIcon[regular]{lock} & 56.9 & 63.8 \\
Qwen3.7-Plus         & API & 397B-A17B & 49.6 & 81.0 \\
UI-Venus-2.0-27B     & Closed & 27B & - & 84.0 \\
UI-Tars-2.0          & Closed & 230B-A23B & - & 73.3 \\
MAI-UI-235B-A22B     & Closed & 235B-A22B & - & 76.7 \\
Xiaomi-GUI-0-30B-A3B & Closed & 30B-A3B & - & 78.9 \\
Step-GUI             & Closed & 8B & - & 80.2 \\
\midrule
\rowcolor{sectionblue}
\multicolumn{5}{l}{\textit{Open Source}} \\
GLM5.3-Flash          & API & 320B-A18B & 66.4 & 78.4 \\
Qwen3.8-Max           & API & 2.4T-A95B & 64.2 & \textbf{85.3} \\
Qwen3.8-Flash         & API & 125B-A6B & 59.1 & 84.5 \\
Qwen3.8-27B           & Open & 27B & 58.0 & 81.9 \\
Qwen3.6-35B-A3B       & Open & 35B-A3B & 40.7 & 73.3 \\
Qwen3.6-27B           & Open & 27B & 64.4 & 70.3 \\
Qwen3.5-122B-A10B     & Open & 122B-A10B & 57.3 & 66.4 \\
Qwen3.5-35B-A3B       & Open & 35B-A3B & 42.2 & 71.1 \\
Qwen3.5-27B           & Open & 27B & 53.7 & 64.2 \\
GUI-Owl-1.5-32B-Think & Open & 32B & 47.1 & 71.6 \\
UI-Venus-2.0-9B       & Open & 9B & 49.6 & 80.2 \\
UI-Venus-1.5-30B-A3B  & Open & 30B-A3B & 24.1 & 77.6 \\
PhoneBuddy            & Open & 4B & 39.0 & 83.2 \\
\midrule
\rowcolor{sectionblue}
\multicolumn{5}{l}{\textit{Ours}} \\
BlueLM-GUI           & Ours & 35B-A3B & \textbf{87.4} & \textbf{84.9} \\
\bottomrule
\end{tabular}
\caption{Comparison of models on \BenchPro and AndroidWorld benchmarks.}
\label{tab:gui_only}
\end{table}

%% file: tables/tab_cpt.tex
\begin{table}[t]
    \centering
    \small
    \begin{tabular}{lccc}
        \toprule
        \textbf{Model} & \textbf{\ScreenZH} & \textbf{\ScreenEN} & \textbf{\BenchBasic} \\
        \midrule
        Base                & 95.48 & \textbf{94.34} & 36.6 \\
        Base + SFT          & \textbf{96.48} & 93.08 & 64.1 \\
        Base + CPT + SFT    & 96.13 & 91.90 & \textbf{70.8} \\
        \bottomrule
    \end{tabular}
    \caption{CPT is necessary. Both variants share one base model and an identical SFT configuration, and differ only in whether CPT precedes SFT.}
    \label{tab:cpt}
\end{table}

%% file: tables/tab_ratio.tex
\begin{table}[t]
    \centering
    \small
    \begin{tabular}{lccc}
        \toprule
        \textbf{Mixture} & \textbf{\ScreenZH} & \textbf{\ScreenEN} & \textbf{\BenchBasic} \\
        \midrule
        Screen-heavy & 96.13 & \textbf{91.90} & 70.8 \\
        Traj-heavy   & \textbf{96.60} & 90.41 & \textbf{71.9} \\
        \bottomrule
    \end{tabular}
    \caption{Cutting screen understanding data and raising the trajectory share improves execution. Traj-heavy uses a CPT corpus 40\% smaller.}
    \label{tab:ratio}
\end{table}

%% file: tables/tab_online.tex
\begin{table}[t]
    \centering
    \small
    \begin{tabular}{llc}
        \toprule
        \textbf{Model} & \textbf{RL setting} & \textbf{\BenchBasic} \\
        \midrule
        SFT           & --                          & 74.9 \\
        + Offline RL  & replay of collected trajectories & 73.9 \\
        + Online RL   & real-device online interaction   & \textbf{82.0} \\
        \bottomrule
    \end{tabular}
    \caption{RL needs a real-device environment. Both RL variants share one SFT starting point.}
    \label{tab:online}
\end{table}

%% file: tables/tab_reward.tex
\begin{table}[t]
    \centering
    \small
    \begin{tabular}{llcc}
        \toprule
        \textbf{Model} & \textbf{Judging} & \textbf{Passes} & \textbf{\BenchPro} \\
        \midrule
        RL-holistic & holistic, disjunctive     & 1 & 72.52 \\
        RL-atomic   & atomic, frame-bound       & 5 & \textbf{77.42} \\
        \bottomrule
    \end{tabular}
    \caption{Reward reliability drives RL. Atomic judging checks conditions that must all hold, each bound to a designated frame; Passes counts judging passes.}
    \label{tab:reward}
\end{table}

%% file: sections/related_works.tex
\section{Related Work}
\label{sec:related}

\subsection*{From Framework Agents to Native GUI Agents}

Early GUI agents are typically constructed as multi-agent orchestration systems built upon general-purpose VLMs. Methods such as the Mobile-Agent series~\citep{mobileagent} and AppAgent~\citep{appagent} compose specialized planner, executor, reflector, and memory modules around foundation models such as GPT-4o~\citep{GPT-4O}, Gemini~\citep{gemini1}, or Qwen2-VL~\citep{qwen2-vl}. The performance of these approaches is largely bounded by the underlying VLM, and inference latency grows with module depth and the number of interaction rounds. Moreover, reflection resides in the runtime procedure rather than being internalized into model weights. To overcome these limitations, recent research has increasingly shifted toward native GUI agents, in which perception, grounding, planning, and execution capabilities are integrated into a single model through post-training. CogAgent~\citep{cogagent}, SeeClick~\citep{seeclick}, OS-Atlas~\citep{screenspotv2}, and AutoGLM~\citep{AutoGLM} established this paradigm, followed by UI-TARS~\citep{uitars}, UI-TARS-1.5~\citep{uitars15}, and UI-TARS-2~\citep{uitars2}, GUI-Owl in Mobile-Agent-v3 and Mobile-Agent-v3.5~\citep{Mobile-agent-v3,Mobile-agent-v3.5}, MAI-UI~\citep{maiui}, Step-GUI~\citep{step-gui}, UI-Venus-1.5~\citep{uivenus15}, PhoneBuddy~\citep{phonebuddy}, Xiaomi GUI-0~\citep{cao2026xiaomi}, and HyMobileAgent~\citep{hymobileagent}. A parallel line extends the paradigm from mobile to general computer use, including MobileIPL~\citep{mobileipl}, GTA1~\citep{gta1}, OpenCUA~\citep{opencua}, UltraCUA~\citep{ultracua}, Fara-7B~\citep{fara7b}, EvoCUA and EvoCUA-1.5~\citep{evocua,evocua15}, CoME~\citep{come}, and ToolCUA~\citep{toolcua}. As native models mature, their capabilities are increasingly determined by the data engines behind them: public corpora such as Android-in-the-Wild~\citep{aitw}, AndroidControl~\citep{androidcontrol}, and AMEX~\citep{amex} provide scale but remain success-biased; automated synthesis pipelines such as AUTO-Explorer~\citep{autoexplorer}, AndroidLab~\citep{androidlab}, GUI-Odyssey~\citep{guiodyssey}, OpenMobile~\citep{openmobile}, and UI-Genie~\citep{uigenie} reduce annotation cost through self-exploration, grounded instruction generation, and self-improving evaluation; and industrial systems have converged on closed-loop data engines---MAI-UI~\citep{maiui} lets agents construct tasks, environments, and failure diagnoses, HyMobileAgent~\citep{hymobileagent} scales a perception flywheel and a video-derived knowledge pipeline over thousands of sandbox and real-device instances, PhoneBuddy~\citep{phonebuddy} shows that mock-app training complements rather than replaces real-app data, and Xiaomi GUI-0~\citep{cao2026xiaomi} recycles rollout failures into corrected actions and recovery demonstrations. Our work follows this native-agent line but adopts a distinct focus on two levels. First, rather than optimizing for static or emulator-dominated benchmarks, we target the constraints of real mobile deployment, including authentic account states, system dialogs, payment authentication, risk-control interventions, network variability, and client-side version drift. Second, whereas existing pipelines largely follow a ``keep-clean-success, discard-failure'' pattern, our data engine treats every sample as a source of supervision: we decouple CPT and SFT pools so that a single real-device collection pass yields both step-level perceptual supervision and trajectory-level planning supervision, and route failed and environment-error trajectories---collected exclusively on real devices so that long-tail conditions are covered---to a unified error-correction and derivation module for step-level correction, retroactive query alignment, and counterfactual derivation.

\subsection*{RL-based GUI Agent Training}

DigiRL~\citep{digirl} is among the first to show that online reinforcement learning can substantially outperform supervised fine-tuning for device-control tasks, underscoring the value of interaction feedback. A wave of R1-style work follows, shifting the paradigm from purely offline imitation learning toward interactive policy optimization: UI-R1~\citep{ui-r1}, GUI-R1~\citep{guir1}, InfiGUI-R1~\citep{infiguir1}, and Mobile-R1~\citep{mobile-r1}. Subsequent efforts scale RL to industrial-grade infrastructure: UI-TARS-2~\citep{uitars2} deploys thousands of parallel virtual machines to support large-scale RL rollouts, MobileRL~\citep{mobilerl} introduces adaptive GRPO under heavy-tailed task difficulty, MAI-UI~\citep{maiui} supports online RL on trajectories exceeding one hundred turns with over ten thousand concurrent environments, and UI-Venus-1.5~\citep{uivenus15} combines a four-stage pipeline of mid-training, offline RL, online RL, and model merging. A complementary line internalizes reasoning and self-correction into the policy itself: ReAct~\citep{react} and Reflexion~\citep{reflexion} establish the primitives of interleaved reasoning, action, and verbal self-critique; UI-TARS~\citep{uitars} integrates a System-2 reasoning module before each action; and Xiaomi GUI-0~\citep{cao2026xiaomi} turns failure trajectories into recovery demonstrations. Together, these studies establish reinforcement learning and built-in recovery as central directions for long-horizon GUI agents. Our training recipe builds on this line but closes the loop differently: rather than treating trajectories as either SFT demonstrations or RL rollouts with a one-way flow from data production to training, we route each trajectory according to its outcome---partially successful rollouts remain in the RL pool as the strongest exploration signal, fully failed rollouts flow to the ECDM and re-enter as SFT data, and only fully mastered queries retire from the pool---so that every trajectory continuously contributes to a co-evolving model--corpus curriculum.

\subsection*{Real-Device and Virtualized GUI Benchmarks}

Existing mobile GUI benchmarks can be broadly divided into virtualized (emulator/sandbox) environments and real-device evaluations. GUI grounding is benchmarked by ScreenSpot~\citep{seeclick}, ScreenSpot-Pro~\citep{screenspotpro}, and MMBench-GUI~\citep{mmbench_gui}, while interactive evaluation is dominated by sandbox environments with programmatic verification: AndroidWorld~\citep{androidworld} provides execution-based grading on emulators, and OSWorld~\citep{osworld} and WebArena~\citep{webarena} play analogous roles for desktop and browser. This sandbox line has since been extended primarily along coverage and task distribution: A3~\citep{a3}, Mobile-Bench-v2~\citep{mobilebenchv2}, and PhoneWorld~\citep{phoneworld} broaden mobile platform coverage, while QwenClawBench~\citep{qwenclawbench} probes real-user task distributions. These settings offer scalable and reproducible evaluation, but their state distributions remain biased toward simplified environments and do not fully capture the account states, page dynamics, and business logic of real applications. To improve fidelity, simulation platforms such as MobileGym~\citep{mobilegym} and SimuWoB~\citep{simuwob} approximate the pages, databases, and partial functions of commercial applications, enabling scalable and verifiable benchmarking while remaining distinct from their real counterparts in state distributions and client-side behaviors. Real-device benchmarks, by contrast, evaluate agents directly on physical devices and live applications: MobileBench-OL~\citep{mobilebench-ol} measures task execution, reasoning, and noise robustness in an online real-device environment; MobileWorld~\citep{mobileworld} and AndroidDaily~\citep{androiddaily} target realistic daily-use scenarios; Xiaomi's RealMobile~\citep{cao2026xiaomi} retains rule-based programmatic evaluation on live apps; and MAI-UI's MobileWorld-Real~\citep{maiui} introduces a multi-judge AutoJudge protocol for trajectory-level evaluation under live conditions. Our benchmark likewise runs on physical devices and live applications, but is designed so that every query is an accountable unit: queries are constructed with explicit functional-coverage indexing against app function trees, annotated with difficulty, ambiguity, and risk metadata, and evaluated with a protocol that separates model failure from environment failure and attributes residual disagreement to human adjudication---turning the benchmark from a static scorecard into a diagnostic instrument whose results feed back into data production and the RL query pool.

%% file: sections/conclusion.tex
\section{Conclusion}
\label{sec:conclusion}

We have presented \OurModel, a 35B-A3B mobile GUI agent built as a real-device-centric flywheel for self-improvement. It targets three structural gaps that block industrial deployment, pairing each with a principle: the \emph{distribution gap} between sandbox training and production is answered by \emph{Every Rollout Is Real}, a three-stage CPT--SFT--RL recipe run end-to-end on hundreds of real phones; the \emph{utilization gap}, in which costly real-device failures are discarded, by \emph{Every Sample Matters}, an ECDM that salvages failed trajectories back into supervision; and the \emph{attribution gap}, in which fixed benchmarks saturate, by \emph{Every Query Evolves}, a quota-driven metric system whose benchmarks are upgraded as the model improves. \OurModel reaches 87.4 on \BenchPro, 5.1 points above the best closed-source API model, and 84.9 on AndroidWorld, competitive with closed-source models that activate far more parameters. Grounding the entire loop---training, salvage, and evaluation---in real devices is what makes the resulting capability transferable to deployment.

Looking ahead, we plan to scale the flywheel to broader device and distributions and to extend the interaction paradigm toward GUI + API + CLI fusion. Of these directions, the one most aligned with the self-evolution thesis is to reduce the human participation the loop still requires, moving it toward greater autonomy.

%% file: sections/contributors.tex
\clearpage
\section{Contributors}

\noindent\textbf{Supervisors}

\vspace{0.5\baselineskip}
\begin{itemize}[leftmargin=1.2em, itemsep=0.1em, parsep=0pt, topsep=0pt]
    \item Xiaohu Ruan$^{\dagger}$
    \item Xiaoxin Chen
\end{itemize}

\vspace{0.1\baselineskip}

\noindent
\begin{minipage}[t]{0.48\textwidth}
\textbf{Contributors}

\vspace{0.5\baselineskip}
\begin{itemize}[leftmargin=1.2em, itemsep=0.1em, parsep=0pt, topsep=0pt]
    \item Tong Ye$^{*}$
    \item Kunyang Han$^{*}$
    \item Guozhi Wang$^{*}$
    \item Longqiang Luo$^{\dagger}$
    \item Zhifeng Ding
    \item Yongxiang Zhang
    \item Xiaolei Shen
    \item Yuxuan Zhang
    \item Zhuping Zhang
    \item Tao Xu
    \item Yue Pan
    \item Yucheng Zhao
    \item Yupei Hu
    \item Yuanjiang Ouyang
    \item Danfeng Shen
    \item Runqi Lin
    \item Hongda Cai
    \item Zhaoxiong Wang
    \item Mengjia Yan
    \item Yingjie Zhong
    \item Chen Zhou
    \item Zeyu Zhang
    \item Xuwen Zhu
\end{itemize}
\end{minipage}%
\hfill
\begin{minipage}[t]{0.48\textwidth}
\vphantom{\textbf{Contributors}}

\vspace{0.6\baselineskip}
\begin{itemize}[leftmargin=1.2em, itemsep=0.1em, parsep=0pt, topsep=0pt]
    \item Penggang Shi
    \item Mingcheng Luo
    \item Ziyang Wu
    \item Min Jin
    \item Mingfu Shen
    \item Zairong Xu
    \item Fan Zhang
    \item Hao Wang
    \item Liang Liu
    \item Zhulin Xie
    \item Lijun Yao
    \item Xiao Liang
    \item Liangmin Wen
    \item Liqiang Feng
    \item Feilong Wu
    \item Min Hu
    \item Min Chen
    \item Guanjing Xiong
\end{itemize}
\end{minipage}

\vspace{2\baselineskip}
\noindent
$^{*}$ Co-first authors\quad
$^{\dagger}$ Corresponding authors

%% file: sections/appendix.tex
\section{Appendix}

\begin{figure}[h]
    \centering
    \includegraphics[width=0.98\linewidth]{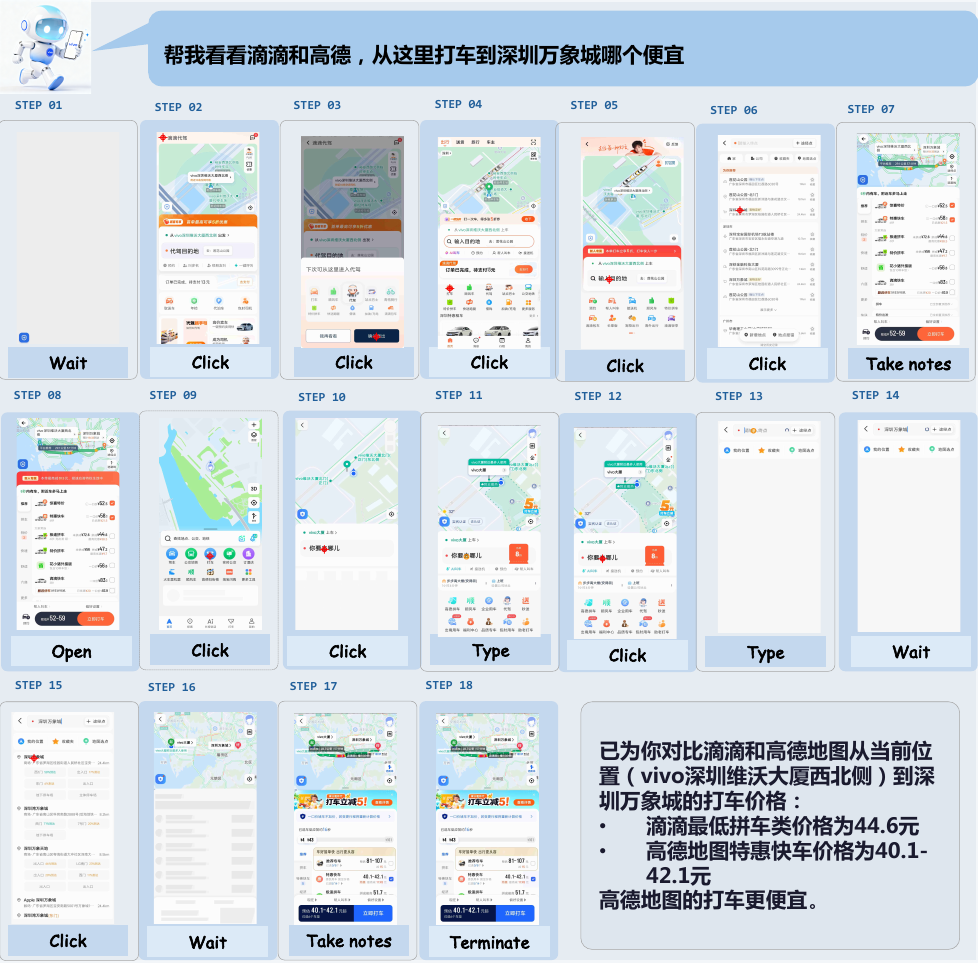}
    \caption{Case. A trajectory of \OurModel comparing ride-hailing fares to Shenzhen MixC across DiDi and Amap.}
    \label{fig:case5}
\end{figure}

\begin{figure}[h]
    \centering
    \includegraphics[width=0.98\linewidth]{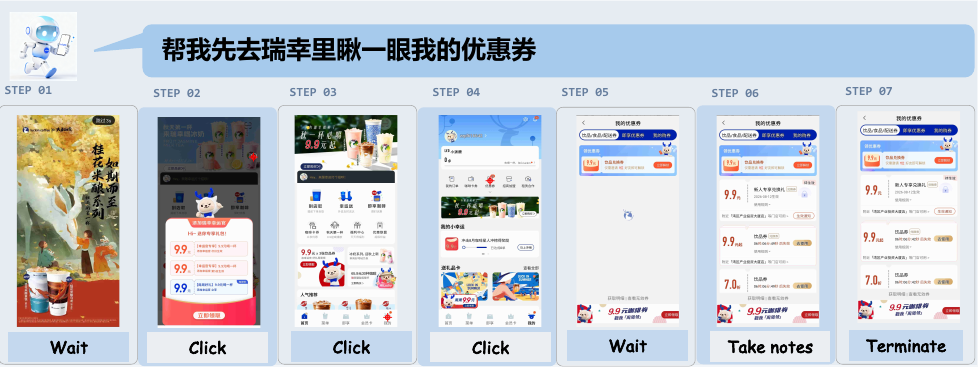}
    \caption{Case. An example of \OurModel completing ``check my coupons'' on Luckin Coffee.}
    \label{fig:case1}
\end{figure}

\begin{figure}[h]
    \centering
    \includegraphics[width=0.98\linewidth]{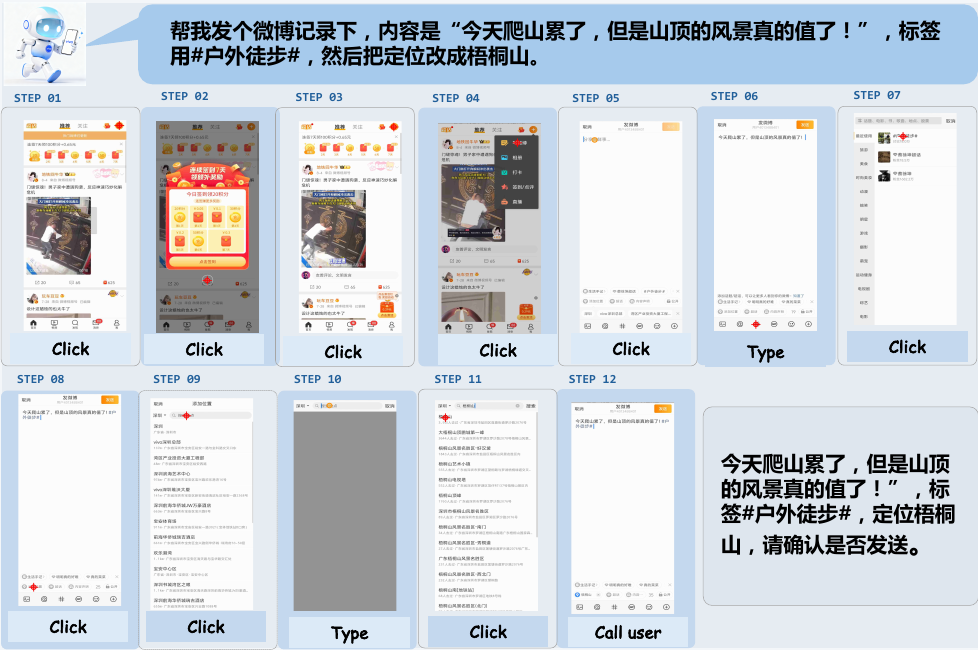}
    \caption{Case. A trajectory of \OurModel posting a Weibo with a custom hashtag and location.}
    \label{fig:case2}
\end{figure}

\begin{figure}[h]
    \centering
    \includegraphics[width=0.94\linewidth]{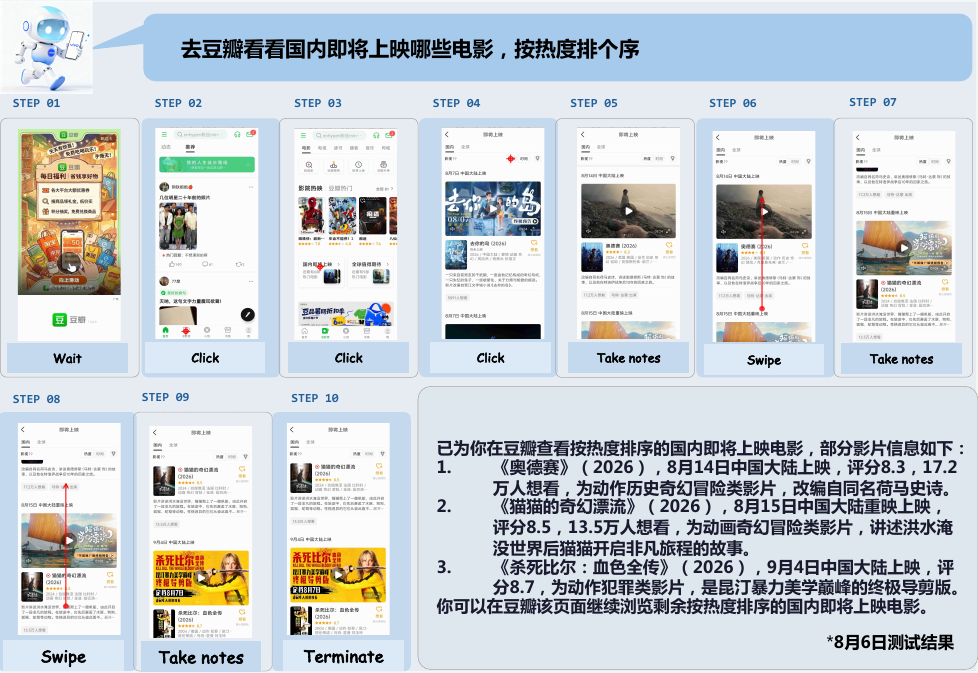}
    \caption{Case. A trajectory of \OurModel browsing upcoming movies on Douban ranked by popularity.}
    \label{fig:case3}
\end{figure}

\begin{figure}[h]
    \centering
    \includegraphics[width=0.94\linewidth]{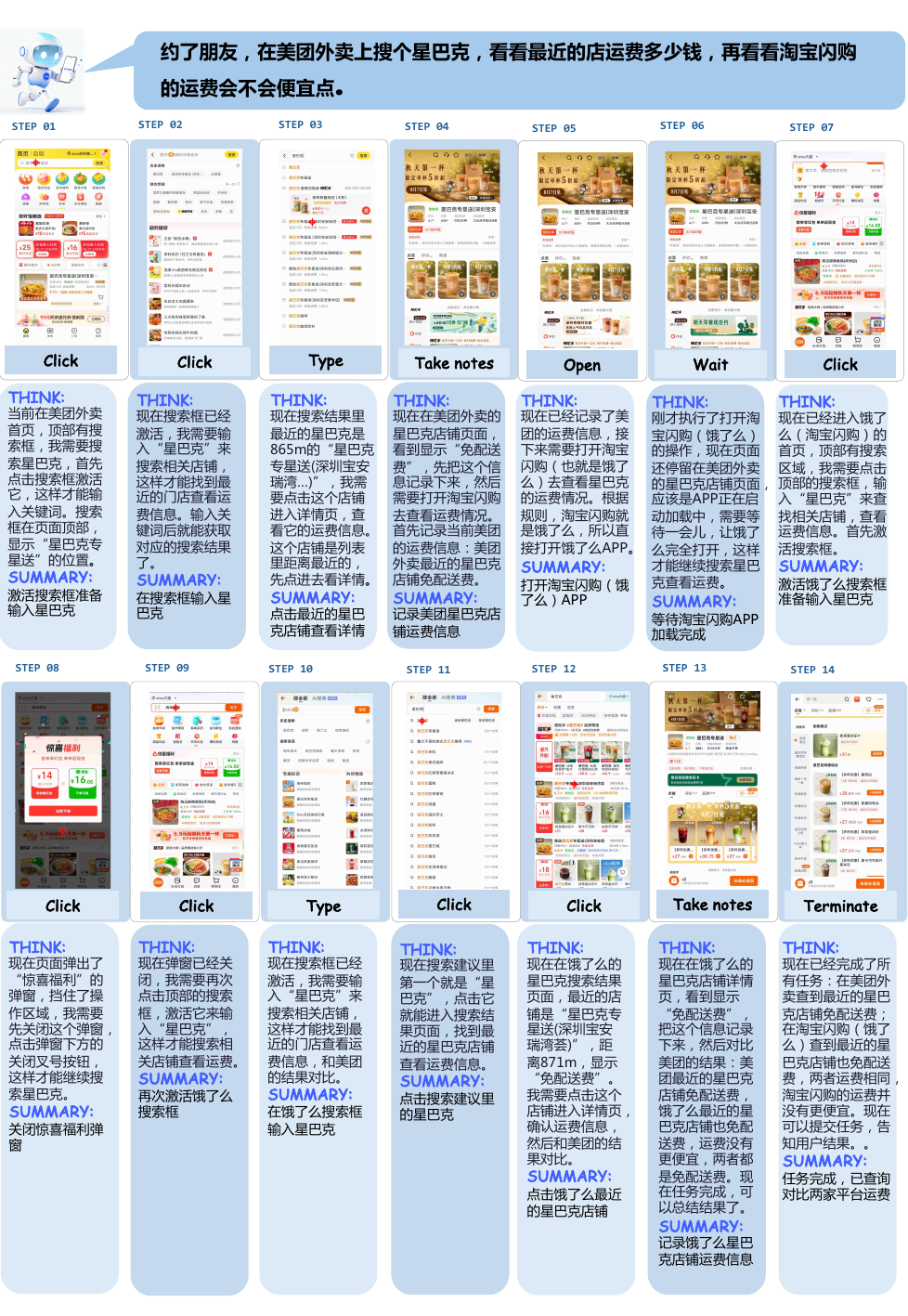}
    \caption{Case. A trajectory of \OurModel comparing Starbucks delivery fees across Meituan and Taobao.}
    \label{fig:case4}
\end{figure}

%% file: references.bib
@inproceedings{zhuo2025bigcodebench,
  title={BigCodeBench: Benchmarking Code Generation with Diverse Function Calls and Complex Instructions},
  author={Terry Yue Zhuo and Vu Minh Chien and Jenny Chim and Han Hu and Wenhao Yu and Ratnadira Widyasari and Imam Nur Bani Yusuf and Haolan Zhan and Junda He and Indraneil Paul and Simon Brunner and Chen GONG and James Hoang and Armel Randy Zebaze and Xiaoheng Hong and Wen-Ding Li and Jean Kaddour and Ming Xu and Zhihan Zhang and Prateek Yadav and Naman Jain and Alex Gu and Zhoujun Cheng and Jiawei Liu and Qian Liu and Zijian Wang and David Lo and Binyuan Hui and Niklas Muennighoff and Daniel Fried and Xiaoning Du and Harm de Vries and Leandro Von Werra},
  booktitle={The Thirteenth International Conference on Learning Representations},
  year={2025},
  url={https://openreview.net/forum?id=YrycTjllL0}
}

@article{maiui,
  title         = {{MAI-UI Technical Report: Real-World Centric Foundation GUI Agents}},
  author        = {Zhou, Hanzhang and Zhang, Xu and Tong, Panrong and Zhang, Jianan and Chen, Liangyu and Kong, Quyu and Cai, Chenglin and Liu, Chen and Wang, Yue and Zhou, Jingren and Hoi, Steven},
  journal       = {arXiv preprint arXiv:2512.22047},
  year          = {2025},
  eprint        = {2512.22047},
  archivePrefix = {arXiv},
  primaryClass  = {cs.CV},
  url           = {https://arxiv.org/abs/2512.22047}
}

@article{hymobileagent,
  title         = {{HyMobileAgent: Data-Environment Co-Scaling for Efficient GUI Agents}},
  author        = {Shen, Huawen and Tang, Zhengyang and Peng, Shangpin and Wu, Liang and Zhang, Anran and Wang, Weinong and Guo, Yiduo and Li, Chenxin and Fang, Zhengyao and Ding, Yang and Li, Junyi and Tang, Fei and Ruan, Zheng and Zhang, Yi and Zhou, Xingran and Yang, Dingchen and Fan, Sunqi and Wan, Zhiyi and Hu, Han and Lai, Xin and Lyu, Pengyuan and Zhang, Chengquan},
  journal       = {arXiv preprint arXiv:2607.14548},
  year          = {2026},
  eprint        = {2607.14548},
  archivePrefix = {arXiv},
  primaryClass  = {cs.CV},
  url           = {https://arxiv.org/abs/2607.14548}
}

@article{phonebuddy,
  title         = {{PhoneBuddy: Training Open Models for Agentic Phone Use with Real and Mock App Environments}},
  author        = {{Tencent Hunyuan Team}},
  journal       = {arXiv preprint arXiv:2606.23049},
  year          = {2026},
  eprint        = {2606.23049},
  archivePrefix = {arXiv},
  url           = {https://arxiv.org/abs/2606.23049}
}

@article{cao2026xiaomi,
  title         = {{Xiaomi-GUI-0 Technical Report}},
  author        = {Cao, Wanxia and Duan, Chengzhen and Fu, Pei and Gao, Pengzhi and Lian, Niu and Liu, Fazhan and Liu, Hui and Qu, Heng and Wu, Qinzhuo and Yu, Zhehao and others},
  journal       = {arXiv preprint arXiv:2606.31410},
  year          = {2026},
  eprint        = {2606.31410},
  archivePrefix = {arXiv}
}

@article{openmobile,
  title         = {{OpenMobile: An Open Data Synthesis Framework for Mobile Agents}},
  author        = {Chen, Kan and Lu, Shengqiao and others},
  journal       = {arXiv preprint arXiv:2604.15093},
  year          = {2026},
  eprint        = {2604.15093},
  archivePrefix = {arXiv},
  url           = {https://arxiv.org/abs/2604.15093}
}

@inproceedings{uigenie,
  title         = {{UI-Genie: A Self-Improving Approach for GUI Agents}},
  author        = {Xiao, Jun and others},
  booktitle     = {Advances in Neural Information Processing Systems},
  year          = {2025},
  eprint        = {2505.21496},
  archivePrefix = {arXiv},
  url           = {https://arxiv.org/abs/2505.21496}
}

@inproceedings{mobileagent,
  title         = {{Mobile-Agent: Autonomous Multi-Modal Mobile Device Agent with Visual Perception}},
  author        = {Wang, Junyang and Xu, Haiyang and Ye, Jiaho and Huang, Ming and Yan, Ming and Zhang, Ji and Huang, Fei and others},
  booktitle     = {arXiv preprint arXiv:2401.16158},
  year          = {2024},
  eprint        = {2401.16158},
  archivePrefix = {arXiv}
}

@inproceedings{appagent,
  title         = {{AppAgent: Multimodal Agents as Smartphone Users}},
  author        = {Yang, Zhao and Zhang, Jiaxuan and others},
  booktitle     = {arXiv preprint arXiv:2312.13771},
  year          = {2023},
  eprint        = {2312.13771},
  archivePrefix = {arXiv}
}

@misc{GPT-4O,
  title         = {{GPT-4o System Card}},
  author        = {{OpenAI}},
  year          = {2024},
  url           = {https://openai.com/index/gpt-4o-system-card/}
}

@misc{gemini1,
  title         = {{Gemini: A Family of Highly Capable Multimodal Models}},
  author        = {{Gemini Team}},
  year          = {2024},
  eprint        = {2312.11805},
  archivePrefix = {arXiv}
}

@article{qwen2-vl,
  title         = {{Qwen2-VL: Enhancing Vision-Language Model's Perception of the World at Any Resolution}},
  author        = {Wang, Peng and Bai, Shuai and Tan, Sinan and Wang, Shijie and Fan, Zhihao and Bai, Jinze and Chen, Keqin and Liu, Xuejing and Wang, Jialin and others},
  journal       = {arXiv preprint arXiv:2409.12191},
  year          = {2024},
  eprint        = {2409.12191},
  archivePrefix = {arXiv}
}

@article{uitars,
  title         = {{UI-TARS: Pioneering Automated GUI Interaction with Native Agents}},
  author        = {Qin, Yujia and Ye, Yining and Fang, Junjie and Wang, Haoming and others},
  journal       = {arXiv preprint arXiv:2501.12326},
  year          = {2025},
  eprint        = {2501.12326},
  archivePrefix = {arXiv}
}

@article{uitars2,
  title         = {{UI-TARS-2 Technical Report: Advancing GUI Agent with Multi-Turn Reinforcement Learning}},
  author        = {{ByteDance Seed}},
  journal       = {arXiv preprint arXiv:2509.02544},
  year          = {2025},
  eprint        = {2509.02544},
  archivePrefix = {arXiv}
}

@article{screenspotv2,
  title         = {{OS-Atlas: A Foundation Action Model for Generalist GUI Agents}},
  author        = {Wu, Zhiyong and Liu, Zhenyu and others},
  journal       = {arXiv preprint arXiv:2410.23218},
  year          = {2024},
  eprint        = {2410.23218},
  archivePrefix = {arXiv}
}

@inproceedings{seeclick,
  title         = {{SeeClick: Harnessing GUI Grounding for Advanced Visual GUI Agents}},
  author        = {Cheng, Kanzhi and Sun, Qiushi and Chu, Yougang and Xu, Fangzhi and Li, YanTao and Zhang, Jianbing and Wu, Zhiyong},
  booktitle     = {Proceedings of the 62nd Annual Meeting of the Association for Computational Linguistics},
  year          = {2024},
  eprint        = {2401.10935},
  archivePrefix = {arXiv}
}

@inproceedings{cogagent,
  title         = {{CogAgent: A Visual Language Model for GUI Agents}},
  author        = {Hong, Wenyi and Wang, Weihan and Lv, Qingsong and Xu, Jiazheng and Yu, Wenmeng and Ji, Junhui and Wang, Yan and Wang, Zihan and Zhang, Yuxiao and Li, Juanzi and others},
  booktitle     = {Proceedings of the IEEE/CVF Conference on Computer Vision and Pattern Recognition},
  year          = {2024},
  eprint        = {2312.08914},
  archivePrefix = {arXiv}
}

@article{Mobile-agent-v3,
  title         = {{Mobile-Agent-v3: Foundamental Agents for GUI Automation}},
  author        = {Ye, Jiabo and Zhang, Xi and Xu, Haiyang and Liu, Haowei and Wang, Junyang and Zhu, Zhaoqing and Zheng, Ziwei and Gao, Feiyu and Cao, Jun and Wu, Ziwei and others},
  journal       = {arXiv preprint arXiv:2508.15144},
  year          = {2025},
  eprint        = {2508.15144},
  archivePrefix = {arXiv}
}

@article{Mobile-agent-v3.5,
  title         = {{Mobile-Agent-v3.5: Multi-platform Fundamental GUI Agents}},
  author        = {Xu, Haiyang and Zhang, Xi and Liu, Haowei and Wang, Junyang and Zhu, Zhaozai and Zhou, Shengjie and Hu, Xuhao and Gao, Feiyu and Cao, Junjie and Wang, Zihua and others},
  journal       = {arXiv preprint arXiv:2602.16855},
  year          = {2026},
  eprint        = {2602.16855},
  archivePrefix = {arXiv}
}

@article{AutoGLM,
  title         = {{AutoGLM: Autonomous Foundation Agents for GUIs}},
  author        = {Liu, Xiao and Qin, Yanda and others},
  journal       = {arXiv preprint arXiv:2411.00820},
  year          = {2024},
  eprint        = {2411.00820},
  archivePrefix = {arXiv}
}

@article{step-gui,
  title         = {{Step-GUI Technical Report}},
  author        = {Yan, Sirui and others},
  journal       = {arXiv preprint arXiv:2601.09661},
  year          = {2025},
  eprint        = {2601.09661},
  archivePrefix = {arXiv}
}

@article{aitw,
  title         = {{Android in the Wild: A Large-Scale Dataset for Android Device Control}},
  author        = {Rawles, Christopher and Li, Yifan and others},
  journal       = {arXiv preprint arXiv:2307.10088},
  year          = {2023},
  eprint        = {2307.10088},
  archivePrefix = {arXiv}
}

@article{androidcontrol,
  title         = {{On the Effects of Data Scale on UI Control Agents}},
  author        = {Li, Wei and Bishop, William and Li, Alice and Rawles, Christopher and Appiah-Adu, Folawiyo and others},
  journal       = {arXiv preprint arXiv:2406.03679},
  year          = {2024},
  eprint        = {2406.03679},
  archivePrefix = {arXiv}
}

@article{amex,
  title         = {{AMEX: Android Multi-annotation Expo Dataset for Mobile GUI Agents}},
  author        = {Chai, Yuxiang and Liu, Shuai and others},
  journal       = {arXiv preprint arXiv:2407.17490},
  year          = {2024},
  eprint        = {2407.17490},
  archivePrefix = {arXiv}
}

@article{autoexplorer,
  title        = {{Auto-Explorer: Automated Data Collection for GUI Agent}},
  author       = {Guo, Xiangwu and Gao, Difei and Shou, Mike Zheng},
  journal      = {arXiv preprint arXiv:2511.06417},
  year         = {2025},
  eprint       = {2511.06417},
  archivePrefix= {arXiv}
}

@article{androidlab,
  title         = {{AndroidLab: Training and Systematic Benchmarking of Android Autonomous Agents}},
  author        = {Xu, Yifan and Liu, Xiao and others},
  journal       = {arXiv preprint arXiv:2410.24024},
  year          = {2024},
  eprint        = {2410.24024},
  archivePrefix = {arXiv}
}

@article{guiodyssey,
  title         = {{GUI-Odyssey: A Comprehensive Dataset for Cross-App GUI Navigation on Mobile Devices}},
  author        = {Lu, Quanfeng and Shao, Wenqi and others},
  journal       = {arXiv preprint arXiv:2406.08451},
  year          = {2024},
  eprint        = {2406.08451},
  archivePrefix = {arXiv}
}

@article{ui-r1,
  title         = {{UI-R1: Enhancing Efficient Action Prediction of GUI Agents by Reinforcement Learning}},
  author        = {Lu, Zhengxi and others},
  journal       = {arXiv preprint arXiv:2503.21620},
  year          = {2025},
  eprint        = {2503.21620},
  archivePrefix = {arXiv}
}

@article{guir1,
  title         = {{GUI-R1: A Generalist R1-Style Vision-Language GUI Agent}},
  author        = {Xia, Yiheng and others},
  journal       = {arXiv preprint arXiv:2504.10458},
  year          = {2025},
  eprint        = {2504.10458},
  archivePrefix = {arXiv}
}

@article{infiguir1,
  title         = {{InfiGUI-R1: Advancing Multimodal GUI Agents from Reactive Actors to Deliberative Reasoners}},
  author        = {Liu, Yuhang and others},
  journal       = {arXiv preprint arXiv:2504.14239},
  year          = {2025},
  eprint        = {2504.14239},
  archivePrefix = {arXiv}
}

@article{mobile-r1,
  title         = {{Mobile-R1: Reinforced Interactive Mobile Agent via Task-Level Reasoning}},
  author        = {Chen, Jingxuan and Li, Zhaoqun and Zhang, Tianyu and Zhu, Qianchao and Li, Haoyuan and Li, Xingao and Zhu, Hanpeng and Liu, Hongxin and Xiao, Zhaopeng and Sun, Xing and Deng, Zhihong},
  journal       = {arXiv preprint arXiv:2506.20332},
  year          = {2025},
  eprint        = {2506.20332},
  archivePrefix = {arXiv}
}

@article{digirl,
  title         = {{DigiRL: Training In-The-Wild Device-Control Agents with Autonomous Reinforcement Learning}},
  author        = {Bai, Hao and Zhou, Yifei and Cemri, Mert and Pan, Jiayi and Suhr, Alane and Levine, Sergey and Kumar, Aviral},
  journal       = {arXiv preprint arXiv:2406.11896},
  year          = {2024},
  eprint        = {2406.11896},
  archivePrefix = {arXiv}
}

@article{uivenus15,
  title         = {{UI-Venus-1.5 Technical Report}},
  author        = {Gao, Zhangchi and others},
  journal       = {arXiv preprint arXiv:2602.09082},
  year          = {2026},
  eprint        = {2602.09082},
  archivePrefix = {arXiv}
}

@article{mobilerl,
  title         = {{MobileRL: Online Reinforcement Learning for Mobile Agents}},
  author        = {Wang, Zecheng and Lin, Zhiqi and Li, Juncheng and Yi, Yongfeng and Gu, Qingxu and Dong, Shuaibing and Wang, Shengwei and He, Zhiyuan and Pang, Ming and Chen, Yihao and others},
  journal       = {arXiv preprint arXiv:2510.14844},
  year          = {2025},
  eprint        = {2510.14844},
  archivePrefix = {arXiv}
}

@inproceedings{react,
  title         = {{ReAct}: Synergizing Reasoning and Acting in Language Models},
  author        = {Yao, Shunyu and Zhao, Jeffrey and Yu, Dian and Du, Nan and Shafran, Izhak and Narasimhan, Karthik and Cao, Yuan},
  booktitle     = {International Conference on Learning Representations},
  year          = {2023}
}

@inproceedings{reflexion,
  title         = {{Reflexion}: Language Agents with Verbal Reinforcement Learning},
  author        = {Shinn, Noah and Cassano, Federico and Gopinath, Ashwin and Narasimhan, Karthik and Yao, Shunyu},
  booktitle     = {Advances in Neural Information Processing Systems},
  year          = {2023}
}

@article{screenspotpro,
  title         = {{ScreenSpot-Pro}: {GUI} Grounding for Professional High-Resolution Computer Use},
  author        = {Li, Kaixin and Hu, Ziyang and others},
  journal       = {arXiv preprint arXiv:2504.07981},
  year          = {2025},
  eprint        = {2504.07981},
  archivePrefix = {arXiv}
}

@article{mmbench_gui,
  title         = {{MMBench-GUI}: Hierarchical Multi-Platform Evaluation Framework for {GUI} Agents},
  author        = {Qin, Yining and others},
  journal       = {arXiv preprint arXiv:2507.19448},
  year          = {2025},
  eprint        = {2507.19448},
  archivePrefix = {arXiv}
}

@inproceedings{androidworld,
  title         = {{AndroidWorld}: A Dynamic Benchmarking Environment for Autonomous Agents},
  author        = {Rawles, Christopher and Li, Yifan and others},
  booktitle     = {International Conference on Learning Representations},
  year          = {2025},
  eprint        = {2405.14573},
  archivePrefix = {arXiv}
}

@inproceedings{osworld,
  title         = {{OSWorld}: Benchmarking Multimodal Agents for Open-Ended Tasks in Real Computer Environments},
  author        = {Xie, Tianbao and Zhang, Danyang and Chen, Jixuan and Li, Xiaochuan and Zhao, Siheng and Cao, Ruisheng and others},
  booktitle     = {Advances in Neural Information Processing Systems},
  year          = {2024}
}

@inproceedings{webarena,
  title         = {{WebArena}: A Realistic Web Environment for Building Autonomous Agents},
  author        = {Zhou, Shuyan and Xu, Frank F. and Zhu, Hao and Zhou, Xuhui and Lo, Robert and Sridhar, Abishek and Cheng, Xianyi and others},
  booktitle     = {International Conference on Learning Representations},
  year          = {2024}
}

@article{mobilegym,
  title        = {{MobileGym: A Verifiable and Highly Parallel Simulation Platform for Mobile GUI Agent Research}},
  author       = {Wu, Dingbang and Hao, Rui and Wang, Haiyang and Wu, Shuzhe and Xiao, Han and Li, Zhenghong and Zhou, Bojiang and Ju, Zheng and Liu, Zichen and Fan, Lue and others},
  journal      = {arXiv preprint arXiv:2605.26114},
  year         = {2026},
  eprint       = {2605.26114},
  archivePrefix= {arXiv}
}

@article{simuwob,
  title        = {{SimuWoB: Simulating Real-World Mobile Apps for Fast and Faithful GUI Agent Benchmarking}},
  author       = {Liu, Guohong and Ye, Jialei and Gao, Pengzhi and Liu, Wei and Luan, Jian and Liu, Yunxin and Li, Yuanchun},
  journal      = {arXiv preprint arXiv:2605.25160},
  year         = {2026},
  eprint       = {2605.25160},
  archivePrefix= {arXiv}
}

@article{mobilebench-ol,
  title         = {{MobileBench-OL}: Online Real-Device Benchmarking for Mobile Agents},
  author        = {Zhao, Ziyu and Zhang, Lele and Liu, Yibo and Chen, Shihao and Zhu, Yijie and Chen, Yuxiang and Li, Jiahao and Yang, Zhiyao and Liu, Kai and Yuan, Chunfeng and Huang, Yihua},
  journal       = {arXiv preprint arXiv:2510.22014},
  year          = {2025},
  eprint        = {2510.22014},
  archivePrefix = {arXiv}
}

@article{androiddaily,
  title        = {{AndroidDaily: A Verifiable Benchmark for Mobile GUI Agents on Real-World Closed-Source Applications}},
  author       = {Sui, Yifan and Huang, Xin and Li, Hongbing and Xu, Fang and Lv, Jiahe and Yan, Haolong and Shen, Yeqing and Liu, Litao and Fan, Zhimin and Meng, Ziyang and others},
  journal      = {arXiv preprint arXiv:2605.27761},
  year         = {2026},
  eprint       = {2605.27761},
  archivePrefix= {arXiv}
}

@article{mobileworld,
  title        = {{MobileWorld: Benchmarking Autonomous Mobile Agents in Agent-User Interactive and MCP-Augmented Environments}},
  author       = {Kong, Quyu and Zhang, Xu and Yang, Zhenyu and Gao, Nolan and Liu, Chen and Tong, Panrong and Cai, Chenglin and Zhou, Hanzhang and Zhang, Jianan and Chen, Liangyu and others},
  journal      = {arXiv preprint arXiv:2512.19432},
  year         = {2026},
  eprint       = {2512.19432},
  archivePrefix= {arXiv}
}

@article{opencua,
  title={Opencua: Open foundations for computer-use agents},
  author={Wang, Xinyuan and Wang, Bowen and Lu, Dunjie and Yang, Junlin and Xie, Tianbao and Wang, Junli and Deng, Jiaqi and Guo, Xiaole and Xu, Yiheng and Wu, Chen and others},
  journal={Advances in Neural Information Processing Systems},
  volume={38},
  pages={139756--139806},
  year={2026}
}

@article{fara7b,
  title={Fara-7B: An Efficient Agentic Model for Computer Use},
  author={Awadallah, Ahmed and Lara, Yash and Magazine, Raghav and Mozannar, Hussein and Nambi, Akshay and Pandya, Yash and Rajeswaran, Aravind and Rosset, Corby and Taymanov, Alexey and Vineet, Vibhav and others},
  journal={arXiv preprint arXiv:2511.19663},
  eprint={2511.19663},
  archivePrefix={arXiv},
  year={2025}
}

@article{evocua,
  title={Evocua: Evolving computer use agents via learning from scalable synthetic experience},
  author={Xue, Taofeng and Peng, Chong and Huang, Mianqiu and Guo, Linsen and Han, Tiancheng and Wang, Haozhe and Wang, Jianing and Zhang, Xiaocheng and Yang, Xin and Zhao, Dengchang and others},
  journal={arXiv preprint arXiv:2601.15876},
  eprint={2601.15876},
  archivePrefix={arXiv},
  year={2026}
}

@article{phoneworld,
  title={PhoneWorld: Scaling Phone-Use Agent Environments},
  author={Tang, Zhengyang and Liu, Yuxuan and Lai, Xin and Li, Junyi and Lyu, Pengyuan and Guo, Yiduo and Fang, Zhengyao and Ding, Yang and Zhang, Yi and Wang, Weinong and others},
  journal={arXiv preprint arXiv:2605.29486},
  eprint={2605.29486},
  archivePrefix={arXiv},
  year={2026}
}

@article{ultracua,
  title={Ultracua: A foundation model for computer use agents with hybrid action},
  author={Yang, Yuhao and Yang, Zhen and Dou, Zi-Yi and Nguyen, Anh and You, Keen and Attia, Omar and Szot, Andrew and Feng, Michael and Ramrakhya, Ram and Toshev, Alexander and others},
  journal={arXiv preprint arXiv:2510.17790},
  eprint={2510.17790},
  archivePrefix={arXiv},
  year={2025}
}

@article{evocua15,
  title={EvoCUA-1.5: Online Reinforcement Learning for Multi-turn Computer-Use Agents},
  author={Huang, Mianqiu and Xue, Taofeng and Peng, Chong and Ding, Jinrui and Fan, Sicheng and Hong, Jiale and Gao, Yufei and Zhang, Xiaocheng and Guo, Linsen and Yang, Xin and others},
  journal={arXiv preprint arXiv:2607.09773},
  eprint={2607.09773},
  archivePrefix={arXiv},
  year={2026}
}

@article{a3,
  title={A3: Android agent arena for mobile gui agents},
  author={Chai, Yuxiang and Li, Hanhao and Zhang, Jiayu and Liu, Liang and Liu, Guangyi and Wang, Guozhi and Ren, Shuai and Huang, Siyuan and Li, Hongsheng},
  journal={arXiv preprint arXiv:2501.01149},
  eprint={2501.01149},
  archivePrefix={arXiv},
  year={2025}
}

@article{mobilebenchv2,
  title={Mobile-Bench-v2: A More Realistic and Comprehensive Benchmark for VLM-Based Mobile Agents},
  author={Xu, Weikai and Jiang, Zhizheng and Liu, Yuxuan and Gao, Pengzhi and Liu, Wei and Luan, Jian and Li, Yuanchun and Liu, Yunxin and Wang, Bin and An, Bo},
  journal={arXiv preprint arXiv:2505.11891},
  eprint={2505.11891},
  archivePrefix={arXiv},
  year={2025},
  url={https://arxiv.org/abs/2505.11891}
}

@article{mobileipl,
  title         = {{MobileIPL}: Enhancing Mobile Agents Thinking Process via Iterative Preference Learning},
  author        = {Huang, Kun and Xu, Weikai and Liu, Yuxuan and Wang, Quandong and Gao, Pengzhi and Liu, Wei and Luan, Jian and Wang, Bin and An, Bo},
  journal       = {arXiv preprint arXiv:2505.12299},
  eprint        = {2505.12299},
  archivePrefix = {arXiv},
  year          = {2025}
}

@inproceedings{backtrackagent,
  title         = {{BacktrackAgent}: Enhancing {GUI} Agent with Error Detection and Backtracking Mechanism},
  author        = {Wu, Qinzhuo and Gao, Pengzhi and Liu, Wei and Luan, Jian},
  booktitle     = {Proceedings of the 2025 Conference on Empirical Methods in Natural Language Processing},
  pages         = {4250--4272},
  year          = {2025}
}

@article{qwenuiagent,
  title         = {{Qwen-UI-Agent Technical Report: Real-World Centric Foundation GUI Agents}},
  author        = {{MAI-UI Team}},
  journal       = {arXiv preprint arXiv:2607.28227},
  year          = {2026},
  eprint        = {2607.28227},
  archivePrefix = {arXiv},
  url           = {https://arxiv.org/abs/2607.28227}
}

@article{phoneharness,
  title        = {{PhoneHarness: Harnessing Phone-Use Agents through Mixed GUI, CLI, and Tool Actions}},
  author       = {Li, Chenxin and Fang, Zhengyao and Tang, Zhengyang and Lyu, Pengyuan and Zhou, Xingran and Lai, Xin and Tang, Fei and Wu, Liang and Guo, Yiduo and Wang, Weinong and others},
  journal      = {arXiv preprint arXiv:2606.14832},
  year         = {2026},
  eprint       = {2606.14832}
}

@article{workflowgym,
  title        = {{Workflow-GYM: Towards Long-Horizon Evaluation of Computer-Use Agentic Tasks in Real-World Professional Fields}},
  author       = {Zhu, Liya and Ding, Jingzhe and Zhang, Jian and Xue, Jianbo and Liang, Shihao and Zhang, Ge and Gao, Xiang and Gu, Qingshui and others},
  journal      = {arXiv preprint arXiv:2606.11042},
  year         = {2026},
  eprint       = {2606.11042}
}

@article{terminalbench,
  title        = {{Terminal-Bench: Benchmarking Agents on Hard, Realistic Tasks in Command Line Interfaces}},
  author       = {Merrill, Mike A. and Shaw, Alexander G. and Carlini, Nicholas and Li, Boxuan and Raj, Harsh and Bercovich, Ivan and Shi, Lin and Shin, Jeong Yeon and Walshe, Thomas and others},
  journal      = {arXiv preprint arXiv:2601.11868},
  year         = {2026},
  eprint       = {2601.11868}
}

@article{claweval,
  title        = {{Claw-Eval: Towards Trustworthy Evaluation of Autonomous Agents}},
  author       = {Ye, Bowen and Li, Rang and Yang, Qibin and Liu, Yuanxin and Yao, Linli and Lv, Hanglong and Xie, Zhihui and An, Chenxin and Li, Lei and Kong, Lingpeng and others},
  journal      = {arXiv preprint arXiv:2604.06132},
  year         = {2026},
  eprint       = {2604.06132}
}

@misc{qwenclawbench,
  title        = {{QwenClawBench: Real-User-Distribution Benchmark for OpenClaw Agents}},
  author       = {{Qwen Team} and {Data Team, Alibaba Group}},
  year         = {2026},
  howpublished= {\url{https://github.com/SKYLENAGE-AI/QwenClawBench}}
}

@misc{uitars15,
  title        = {{UI-TARS-1.5}},
  author       = {{ByteDance Seed}},
  year         = {2025},
  howpublished= {\url{https://seed-tars.com/1.5}}
}

@article{gta1,
  title        = {{GTA1: GUI Test-time Scaling Agent}},
  author       = {Yang, Yan and Li, Dongxu and Dai, Yutong and Yang, Yuhao and Luo, Ziyang and Zhao, Zirui and Hu, Zhiyuan and Huang, Junzhe and Saha, Amrita and Chen, Zeyuan and others},
  journal      = {arXiv preprint arXiv:2507.05791},
  year         = {2025},
  eprint       = {2507.05791}
}

@article{toolcua,
  title        = {{ToolCUA: Towards Optimal GUI-Tool Path Orchestration for Computer Use Agents}},
  author       = {Hu, Xuhao and Zhang, Xi and Xu, Haiyang and Qiao, Kyle and Yang, Jingyi and Huang, Xuanjing and Shao, Jing and Yan, Ming and Ye, Jieping},
  journal      = {arXiv preprint arXiv:2605.12481},
  year         = {2026},
  eprint       = {2605.12481}
}

@article{come,
  title        = {{CoME: Empowering Channel-of-Mobile-Experts with Informative Hybrid-Capabilities Reasoning}},
  author       = {Liu, Yuxuan and Xu, Weikai and Huang, Kun and Chen, Changyu and Zhao, Jiankun and Gao, Pengzhi and Liu, Wei and Luan, Jian and Shang, Shuo and Du, Bo and others},
  journal      = {arXiv preprint arXiv:2602.24142},
  year         = {2026},
  eprint       = {2602.24142}
}

@article{singlerollout,
  title        = {{Single-Rollout Asynchronous Optimization for Agentic Reinforcement Learning}},
  author       = {Hou, Zhenyu and Li, Yujiang and Tang, Jie and Dong, Yuxiao},
  journal      = {arXiv preprint arXiv:2607.07508},
  year         = {2026},
  eprint       = {2607.07508}
}

@article{uimopd,
  title        = {{UI-MOPD: Multi-Platform On-Policy Distillation for Continual GUI Agent Learning}},
  author       = {Lian, Niu and Chen, Alan and Yu, Zhehao and Duan, Chengzhen and Liu, Fazhan and Liu, Hui and Fu, Pei and Luan, Jian and Wang, Yaowei and Xia, Shu-Tao and Wang, Jinpeng},
  journal      = {arXiv preprint arXiv:2607.04425},
  year         = {2026},
  eprint       = {2607.04425}
}

@article{openclawrl,
  title        = {{OpenClaw-RL: Train Any Agent Simply by Talking}},
  author       = {Wang, Yinjie and Chen, Xuyang and Jin, Xiaolong and Wang, Mengdi and Yang, Ling},
  journal      = {arXiv preprint arXiv:2603.10165},
  year         = {2026},
  eprint       = {2603.10165}
}

@misc{gemini31pro,
  title        = {{Gemini 3.1 Pro Model Card}},
  author       = {{Google DeepMind}},
  year         = {2026},
  howpublished= {\url{https://deepmind.google/models/model-cards/gemini-3-1-pro/}}
}

@misc{gpt55,
  title        = {{GPT-5.5 System Card}},
  author       = {{OpenAI}},
  year         = {2026},
  howpublished= {\url{https://deploymentsafety.openai.com/gpt-5-5}}
}

@misc{claudeopus48,
  title        = {{Introducing Claude Opus 4.8}},
  author       = {{Anthropic}},
  year         = {2026},
  howpublished= {\url{https://www.anthropic.com/news/claude-opus-4-8}}
}

@article{deepseek-v4,
  title         = {{DeepSeek-V4: Towards Highly Efficient Million-Token Context Intelligence}},
  author        = {DeepSeek-AI and others},
  journal       = {arXiv preprint arXiv:2606.19348},
  year          = {2026},
  eprint        = {2606.19348},
  archivePrefix = {arXiv},
  url           = {https://arxiv.org/abs/2606.19348}
}

@article{kimik3,
  title         = {{Kimi K3: Open Frontier Intelligence}},
  author        = {{Kimi Team} and others},
  journal       = {arXiv preprint arXiv:2607.24653},
  year          = {2026},
  eprint        = {2607.24653},
  archivePrefix = {arXiv},
  url           = {https://arxiv.org/abs/2607.24653}
}
